\documentclass[journal]{IEEEtran}

\usepackage[T1]{fontenc}
\usepackage{cite}
\usepackage{amsmath,amssymb,amsfonts,amsthm,mathtools,bm}
\usepackage{graphicx}
\usepackage[caption=false,font=footnotesize]{subfig}
\usepackage{multirow,array,tabularx,makecell}
\usepackage{booktabs}
\usepackage{algorithm}
\usepackage{algpseudocode}
\usepackage{xcolor}
\usepackage{url}
\usepackage{balance}
\usepackage{hyperref}

\hypersetup{hidelinks}

\DeclareMathOperator*{\argmin}{arg\,min}

\newcolumntype{Y}{>{\centering\arraybackslash}X}
\newcommand{\method}{Chronos}
\newcommand{\R}{\mathbb{R}}
\newcommand{\E}{\mathbb{E}}
\newcommand{\dd}{\mathrm{d}}
\newcommand{\bs}[1]{\boldsymbol{#1}}

\begin{document}

\title{\method: A Physics-Informed Full-History Framework for Non-Markovian Long-Horizon Manipulation}

\author{Yulin Zhou, Yimeng Wang, Nengyu Wang, Shaojia Xing, Shiyun Tu, Xiang Li, Jingkai Zhang, Ningbo Jiang, Yuankai Lin, Hua Yang, Xiangrui Zeng, and Zhouping Yin%
\thanks{Yulin Zhou, Yimeng Wang, Nengyu Wang, Shaojia Xing, Shiyun Tu, Xiang Li, Jingkai Zhang, Ningbo Jiang, Yuankai Lin, Hua Yang, Xiangrui Zeng, and Zhouping Yin are with the School of Mechanical Science and Engineering, Huazhong University of Science and Technology, Wuhan, China.}%
\thanks{Corresponding author: Hua Yang (e-mail: huayang@hust.edu.cn).}%
\thanks{Project page and code: \url{https://github.com/yulinzhouZYL/Chronos}.}%
}

\markboth{IEEE Transactions on Robotics, Under Review}%
{Anonymous Submission: \method}

\maketitle

\begin{abstract}
General-purpose robot policies should be modeled as dynamical systems, yet many current vision-language-action (VLA) and generative imitation learning policies still rely on present observations or short windows. This Markovian shortcut fails in memory-dependent manipulation: identical observations can demand different actions after different histories. We present \method, a physics-informed full-history framework for non-Markovian long-horizon manipulation. The key idea is to elevate observation history from auxiliary context to the latent state of the policy dynamics. At each physical control step, \method\ forms one state-representative token by fusing the current observation and proprioception, so the token sequence is aligned one-to-one with physical time. A selective state space model propagates this causal historical state, which conditions a multimodal coarse action prior through implicit maximum likelihood estimation (IMLE). This prior is then refined by a second-order Schr\"odinger-inspired bridge that predicts acceleration fields, yielding smoother and more physically grounded robot motion. Across 16 simulated tasks and 4 real-world experiments, \method\ is evaluated on precision insertion, general manipulation, and memory-dependent long-horizon control. On RMBench, where success requires remembering task phase, \method\ achieves a 73.6\% average success rate, outperforming the strong Markovian VLA baseline \(\pi_{0.5}\) by \(+62.4\) absolute percentage points, a \(6.6\times\) relative gain, while using \(10\times\) fewer parameters. It also surpasses the state-of-the-art memory VLA Mem-0 by 22.8 points while using over \(30\times\) fewer parameters. In real-world dual-arm experiments using a single RGB camera, \method\ achieves 78\% average success over four tasks, including 72\% on the three memory-dependent tasks, whereas \(\pi_{0.5}\) achieves only 7\% overall and 0\% on the memory-dependent subset. These results suggest that history should not be treated as auxiliary context, but as the latent state of the manipulation policy. Project page and code are available at \url{https://github.com/yulinzhouZYL/Chronos}.
\end{abstract}

\begin{IEEEkeywords}
Imitation learning, non-Markovian policies, state space models, Mamba, Schr\"odinger equation.
\end{IEEEkeywords}

\begin{figure}[!t]
    \centering
    \includegraphics[width=1\columnwidth]{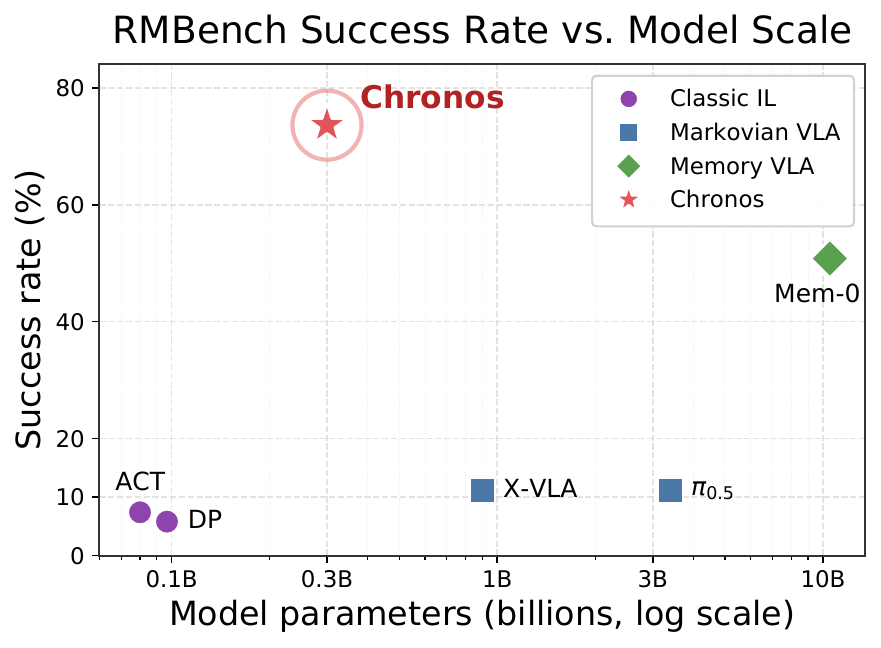}
    \vspace{-1mm}
    \caption{
    RMBench success rate versus model scale.
    Chronos achieves the highest success rate with a compact 0.3B model, outperforming both compact imitation-learning baselines and much larger VLA-based policies.
    }
    \label{fig:rmbench_scale}
    \vspace{-3mm}
\end{figure}

\begin{figure*}[t]
    \centering
    \includegraphics[width=1\textwidth]{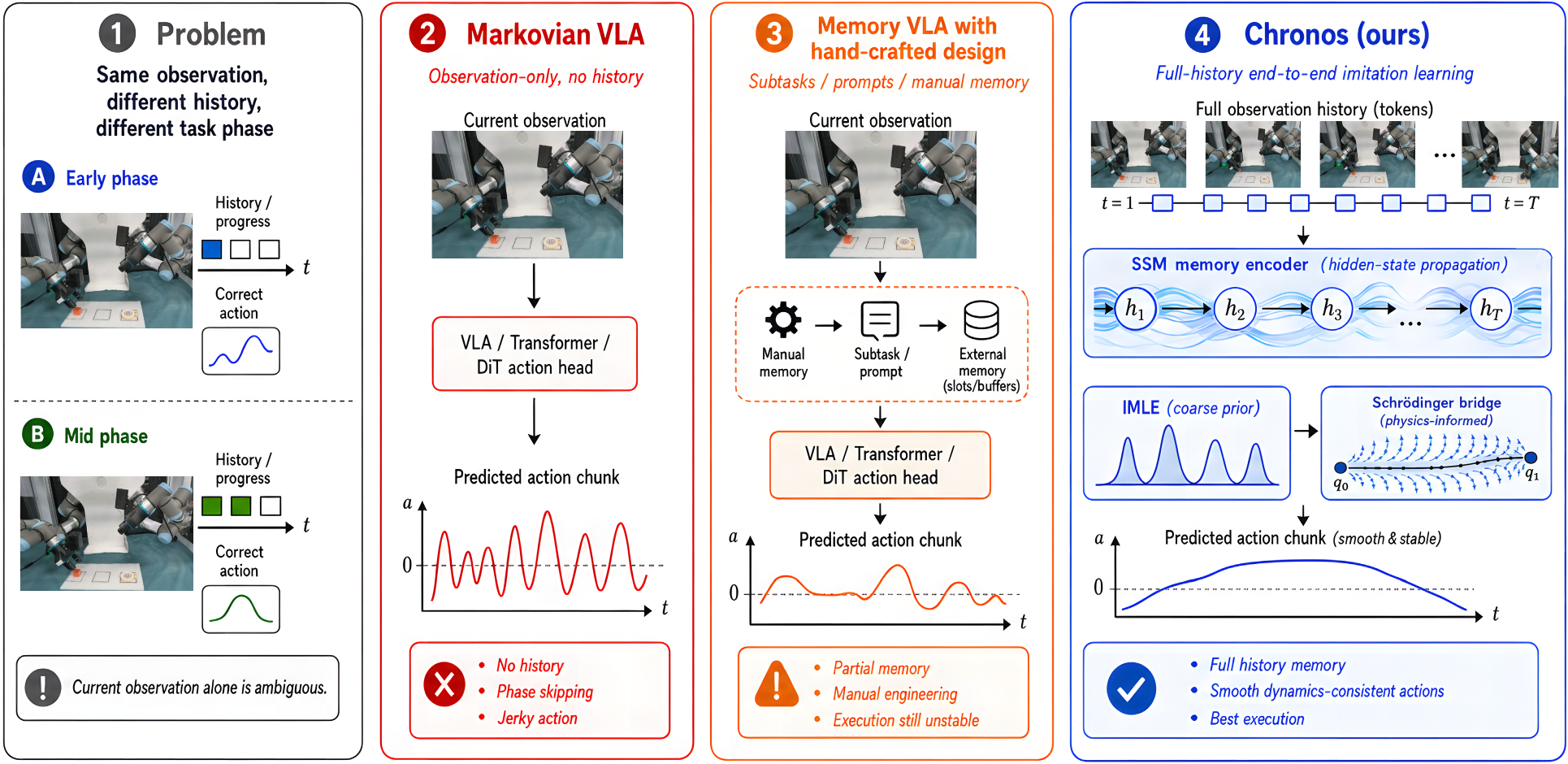}
    \vspace{-1mm}
    \caption{
    Illustration of the core challenge and three policy paradigms.
    Given similar observations but different histories and task phases, Markov policies and hand-crafted memory-based VLA pipelines can fail to produce phase-consistent actions, whereas Chronos uses full-history encoding and a second-order Schr\"odinger-inspired refinement module to generate memory-aware action trajectories.
    }
    \label{fig:teaser}
    \vspace{-2mm}
\end{figure*}

\section{Introduction}

Robotic manipulation is increasingly shaped by large-scale visuomotor foundation models. Robot transformers and VLA systems such as OpenVLA, \(\pi_0\), \(\pi_{0.5}\), RDT-1B, and X-VLA demonstrate that scaling visual, linguistic, and robotic data can yield broadly capable policies \cite{openx,openvla,pi0,pi05,rdt1b,xvla}. More recent systems, including \(\pi_{0.6}\), \(\pi_{0.7}\), RDT-2, and Motus, continue this trajectory toward web-scale robot action modeling and whole-action manipulation (WAM) policies \cite{pi06,pi07,rdt2,motus}. Yet scaling alone does not remove a fundamental ambiguity in robotic control: the present observation is not always a sufficient statistic for the future action.

The difficulty is not only practical but structural. Consider a robot that removes a cup, inspects it, and returns it to the original pose. Before removal and after return, the instantaneous image and point cloud may be almost identical; nevertheless, the correct next action is different because the task phase is different. This is an observation-aliasing problem. The environment may be Markovian in the full physical state, but the robot policy acts on a partial observation, so the imitation problem becomes non-Markovian in the observation space. A policy that only observes the present frame must either guess or collapse to a mean behavior. A policy that sees a short window can still fail if the decisive event occurred earlier than the window. Long-horizon manipulation therefore requires a representation in which history is not an auxiliary prompt but the latent state of the control system itself.

This paper proposes \method, a non-Markovian imitation learning framework built around two principles. First, the temporal axis should represent physical time and nothing else. At each physical time step, the current point cloud or image and the proprioceptive state are encoded into one state-representative token. The full token sequence length equals the expert trajectory length \(L\). A selective state space model (SSM), implemented with Mamba-style linear-time recurrence, then propagates latent memory across the full trajectory \cite{mamba,mamba2,s4}. Unlike detached recurrent training, \method\ preserves temporal credit assignment through the full SSM sequence: a late-stage imitation loss can update the representations that governed earlier phase formation.

Second, action generation should respect the second-order nature of continuous robot motion. Diffusion policies learn score fields over noisy actions, and flow-matching policies learn first-order velocity fields \cite{diffusionpolicy,flowmatching,flowpolicy,actionflow}. These are powerful generative paradigms, but they do not explicitly model acceleration, the quantity that mediates smooth changes of velocity and displacement. \method\ instead introduces a second-order Schr\"odinger-inspired bridge. A conditional implicit generator first samples a coarse multimodal action prior; the bridge then refines that prior by integrating a learned acceleration field in action space. The derivation starts from the Schr\"odinger equation, applies the Madelung transformation, obtains a quantum Hamilton--Jacobi equation and continuity equation, introduces a Gaussian bridge approximation and a Kostin dissipative potential, and yields a stable acceleration-matching loss.

The resulting system combines long-horizon memory and physics-informed action smoothness. Our contributions are threefold:
\begin{enumerate}
    \item We formulate long-horizon imitation learning as full-history state-token modeling. A memory-aware perception--state-space pipeline decouples high-dimensional encoding from temporal state learning, allowing complete trajectory-level credit assignment over one token per physical step.
    \item We propose a physics-informed second-order Schr\"odinger-inspired theory for robot action generation. Instead of treating actions as samples to be denoised or transported by velocity fields, we model action chunks as generalized coordinates refined by acceleration fields, providing a smoother and more physically grounded generative paradigm for robot motion.
    \item We validate \method\ on 16 simulated tasks and 4 real-world experiments. On ALOHA bimanual insertion, controlled ablations show that the second-order Schr\"odinger-inspired bridge improves action smoothness and insertion precision over first-order diffusion and flow-matching action heads. On RoboTwin 2.0, \method\ achieves the best average score among the reported baselines on general manipulation tasks. On RMBench, \method\ achieves a 73.6\% average success rate, outperforming the Markovian VLA baseline \(\pi_{0.5}\) by \(+62.4\) absolute points while using \(10\times\) fewer parameters, and surpassing the memory-aware Mem-0 by 22.8 points while using over \(30\times\) fewer parameters. In real-world dual-arm experiments using a single RGB camera, \method\ achieves 72\% average success on three memory-dependent tasks, whereas \(\pi_{0.5}\) achieves 0\% on the same tasks.

\end{enumerate}

\begin{figure*}[t]
    \centering
    \includegraphics[width=1.0\textwidth]{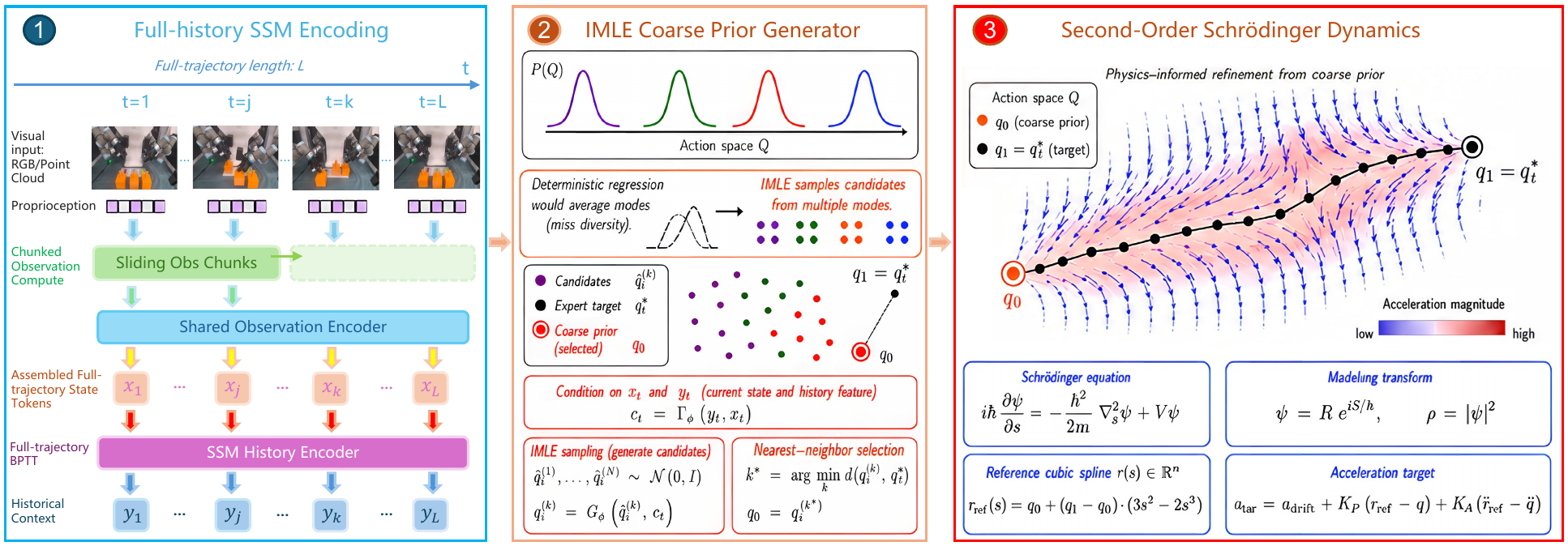}
    \caption{Overview of \method. At each physical time step, Chronos forms one state-representative token \(x_t\) by fusing observation and proprioception. A selective state space model propagates an internal recurrent state \(\bs{h}_t\) and emits a causal historical context \(\bs{y}_t\) over the full trajectory. The historical context conditions a coarse multimodal action prior \(\bs{q}_0\). A second-order Schr\"odinger-inspired bridge then refines \(\bs{q}_0\) into the final action chunk \(\bs{q}_N\) by integrating a learned acceleration field.}
    \label{fig:overview}
\end{figure*}

\section{Related Work}

\subsection{Large VLA and WAM Robot Policies}

Large robot policies have evolved from supervised robot transformers to multimodal foundation models. Open X-Embodiment and OpenVLA further exposed the value of cross-embodiment data and open VLA training \cite{openx,openvla}. The \(\pi_0\) family adopted flow-based action generation for general robot control, with \(\pi_{0.5}\) emphasizing broader deployment and generalization \cite{pi0,pi05}. RDT-1B and X-VLA represent another line of scaled diffusion or transformer-based robot action models, including bimanual and cross-embodiment settings \cite{rdt1b,xvla}. Emerging systems such as \(\pi_{0.6}\), \(\pi_{0.7}\), RDT-2, and Motus point toward even larger WAM/VLA models \cite{pi06,pi07,rdt2,motus}.

These models excel when semantic recognition, broad data coverage, and local geometry are the dominant bottlenecks. However, many are still commonly evaluated or deployed with single-frame, current-state, or short-window conditioning. Our work is complementary to scaling: it asks what dynamical structure should be imposed when task success depends on hidden historical phase rather than only on present visual semantics.

\subsection{Memory-Dependent Robot Manipulation}

Many manipulation tasks are only partially observable from the current sensory frame. The robot may observe a scene that looks almost identical before and after an important historical event, yet the correct next action can be different. This phenomenon has motivated a growing line of memory-centric robotic manipulation benchmarks and policies. RMBench explicitly constructs memory-dependent manipulation tasks and proposes Mem-0 with anchor and sliding memory modules for controlled memory evaluation \cite{rmbench}. RoboMME evaluates temporal, spatial, object, and procedural memory in robotic generalist policies and studies multiple memory-augmented VLA variants \cite{robomme}. MIKASA-Robo introduces a suite of memory-intensive tabletop tasks spanning object, spatial, sequential, and memory-capacity dimensions \cite{mikasa}. LIBERO-Mem focuses on object-level POMDPs, including object motion, object sequence, object relation, and object occlusion, showing that image-based VLA policies struggle when object identity and interaction history must be retained over time \cite{liberomem}. RoboMemArena further expands memory evaluation with long trajectories, memory-related annotations, and paired real-world tasks \cite{robomemarena}.

Recent policies have explored different ways of injecting memory into visuomotor control. MemoryVLA introduces a perceptual-cognitive memory bank for long-horizon robotic manipulation \cite{memoryvla}. ReMem-VLA uses dual-level recurrent queries to carry information across both frames and temporal chunks \cite{rememvla}. MemoAct adopts a hierarchical memory design inspired by the Atkinson--Shiffrin model, separating precise short-term state tracking from compressed long-term retention \cite{memoact}. SAM2Act integrates a visual foundation model with a memory architecture for robotic manipulation \cite{sam2act}. MTIL previously explored full-history imitation learning with state-space memory, but its training relied on detached hidden states under memory constraints, limiting end-to-end temporal credit assignment across the entire trajectory \cite{mtil}. These works establish that memory is not a minor implementation detail but a core capability.

Our thesis is different in form: memory should be the latent state of the policy dynamics itself. Rather than attaching a separate memory system to a Markovian policy or merely extending the observation window, \method\ makes the available past trajectory the native input to a causal linear-time state model. This allows task-phase information to evolve continuously through the policy state and enables late-stage imitation losses to shape earlier phase representations during training.

\subsection{Generative and Physics-Informed Imitation Learning}

Behavior cloning remains the simplest form of imitation learning but suffers from compounding error and mode averaging in multimodal demonstrations \cite{bc,dagger}. Action Chunking with Transformers (ACT) reduces compounding error by predicting multi-step action chunks and has been effective for fine-grained bimanual manipulation \cite{act,aloha}. Diffusion Policy models multimodal action distributions by iterative denoising and remains one of the strongest general-purpose generative imitation baselines \cite{diffusionpolicy}. 3D Diffusion Policy shows that point clouds provide powerful geometric inductive bias for manipulation \cite{dp3}. Point-cloud and equivariant policy learning further improve geometric generalization. Canonical Policy learns canonical 3D representations for SE(3)-equivariant policies, demonstrating strong generalization across object appearance, shape, viewpoint, and robot settings \cite{canonicalpolicy}. Classical point-cloud backbones such as PointNet, PointNet++, and Point Transformer provide the geometric foundation for such 3D policy representations \cite{pointnet,pointnetpp,pointtransformer}. Beyond diffusion, flow matching, rectified flow, action flow matching, and consistency-style policies reduce the sampling burden by learning transport, velocity, or consistency fields \cite{flowmatching,rectifiedflow,flowpolicy,actionflow,consistencypolicy}. IMLE offers another path by learning implicit generators through nearest-sample matching, avoiding the mode-averaging behavior of deterministic regression \cite{imle,imlepolicy}. Most of these methods model direct actions, noise scores, or first-order velocities. \method\ uses IMLE only for the coarse multimodal prior and delegates final refinement to a second-order acceleration field. Schr\"odinger bridges connect stochastic control, entropy-regularized transport, and generative modeling \cite{schrodinger1931,leonard2014,chen2016sb,debortoli2021dsb,vargas2021sb}. Diffusion and flow models can be interpreted within broader probability-flow and transport frameworks \cite{ddpm,scorebased,flowmatching,rectifiedflow}. The Madelung transform rewrites Schr\"odinger dynamics as a continuity equation plus a Hamilton--Jacobi equation with an additional quantum potential \cite{madelung1927,bohm1952,holland1993}. Kostin's dissipative Schr\"odinger equation introduces friction through a nonlinear logarithmic potential while preserving probability conservation \cite{kostin1972}. Symplectic integration has long been used for stable numerical simulation of Hamiltonian systems and has recently influenced neural dynamics and physics-informed learning \cite{symplectic_ode,hamiltonian_nn,neuralode}.

\method\ does not attempt to solve a generic bridge problem over arbitrary high-dimensional distributions. Instead, it builds a control-oriented projection: the action chunk is treated as a generalized coordinate, an implicit generator supplies a coarse start point, the expert action is the endpoint during training, a cubic-spline trajectory supplies the classical guide, and the learned network predicts the acceleration field used to stabilize and refine the bridge.

\section{Problem Formulation}

\subsection{From Markovian Observation Policies to Historical State}

Let an expert demonstration be
\begin{equation}
    \tau=\{(o_t,p_t,a_t^\star)\}_{t=1}^{L},
    \label{eq:traj}
\end{equation}
where \(o_t\) is the exteroceptive observation, \(p_t\) is the proprioceptive state, \(a_t^\star\in\R^{d_a}\) is the expert action, and \(L\) is the actual demonstration length. Conventional behavior cloning often assumes a Markovian observation policy
\begin{equation}
    a_t^\star \approx \pi_\theta(o_t,p_t),
    \label{eq:markov_policy}
\end{equation}
possibly extended to a fixed window
\begin{equation}
    a_t^\star \approx
    \pi_\theta(o_{t-K:t},p_{t-K:t}).
    \label{eq:window_policy}
\end{equation}
Equations~\eqref{eq:markov_policy} and \eqref{eq:window_policy} are valid only when the conditioning variables are sufficient statistics for the expert decision.

Memory-dependent manipulation violates this assumption. There may exist two histories \(\mathcal{H}_t\) and \(\mathcal{H}'_t\) such that
\begin{equation}
\begin{aligned}
    (o_t,p_t)&=(o'_t,p'_t),\\
    \pi^\star(\mathcal{H}_t)&\neq
    \pi^\star(\mathcal{H}'_t),
\end{aligned}
\label{eq:aliasing}
\end{equation}
where
\begin{equation}
    \mathcal{H}_t=\big((o_1,p_1),\ldots,(o_t,p_t)\big).
    \label{eq:history}
\end{equation}
The present observation is then a many-to-one projection of the true task phase. The policy must reconstruct a latent phase variable from the path by which the present was reached.

A useful way to state the issue is through equivalence classes. Define an observation-aliasing relation
\begin{equation}
    \mathcal{H}_t \sim_o \mathcal{H}'_t
    \quad\Longleftrightarrow\quad
    (o_t,p_t)=(o'_t,p'_t).
    \label{eq:alias_relation}
\end{equation}
A Markovian observation policy is well-defined on each equivalence class only if
\begin{equation}
    \pi^\star(\mathcal{H}_t)=\pi^\star(\mathcal{H}'_t)
    \quad
    \forall\ \mathcal{H}_t\sim_o \mathcal{H}'_t .
    \label{eq:markov_sufficient}
\end{equation}
RMBench-style tasks are designed precisely so that \eqref{eq:markov_sufficient} is false. Therefore the correct imitation target is
\begin{equation}
    a_t^\star = \pi^\star(\mathcal{H}_t),
    \label{eq:hist_policy}
\end{equation}
not a function of the current observation alone.

\subsection{Full-History Tokens}

At each physical time step, we map the observation and proprioceptive state to one state-representative token.
\begin{equation}
    \bs{x}_t = \phi_\eta(o_t,p_t)\in\R^d,
    \label{eq:state_token}
\end{equation}
and define the full trajectory token sequence
\begin{equation}
    \bs{X}_{1:L}=(\bs{x}_1,\bs{x}_2,\ldots,\bs{x}_L).
    \label{eq:token_sequence}
\end{equation}
The sequence length in \eqref{eq:token_sequence} is not the number of image patches and not a truncated context size; it is the physical trajectory length. This distinction matters because it prevents the model from mixing spatial token order with temporal order. Temporal modeling capacity is used for memory, not for within-frame bookkeeping.

The history encoder maintains an internal recurrent state and emits a causal historical context:
\begin{equation}
    (\bs{h}_{1:L},\bs{y}_{1:L})
    =
    \mathrm{SSM}_\theta(\bs{X}_{1:L}),
    \qquad
    \bs{h}_t\in\R^{d_h},\quad \bs{y}_t\in\R^{d}.
    \label{eq:ssm_hist}
\end{equation}
Here \(\bs{h}_t\) denotes the internal recurrent state of the selective state space model, while \(\bs{y}_t\) denotes the emitted historical context used for action generation. In forward time, \(\bs{y}_t\) summarizes the available past through the recurrent state \(\bs{h}_t\). During training, losses from all time steps backpropagate through the complete SSM computation graph, enabling long-range temporal credit assignment.

\subsection{Action Chunks as Generalized Coordinates}

For action generation, we use an \(H\)-step action chunk
\begin{equation}
    \bs{q}\in\R^{D},\qquad D=H d_a,
    \label{eq:q_dim}
\end{equation}
as a generalized coordinate in action space. We introduce a bridge time \(s\in[0,1]\) and an auxiliary bridge velocity
\begin{equation}
    \bs{v}(s)=\frac{\dd \bs{q}(s)}{\dd s}.
    \label{eq:bridge_velocity}
\end{equation}
The velocity in \eqref{eq:bridge_velocity} is a derivative in generative bridge time, not necessarily the physical joint velocity.

The proprioceptive state \(p_t\) is not injected as a separate action-generation input. Instead, it is already fused into the state token
\begin{equation}
    \bs{x}_t=\phi_\eta(o_t,p_t).
    \label{eq:state_token_with_prop}
\end{equation}
The IMLE prior is therefore conditioned on the emitted historical context and the current fused state token:
\begin{equation}
    \bs{c}_t = \Gamma_\omega(\bs{y}_t,\bs{x}_t).
    \label{eq:condition_yx}
\end{equation}
The desired conditional second-order dynamics are
\begin{equation}
\begin{aligned}
    \frac{\dd \bs{q}}{\dd s} &= \bs{v},\\
    \frac{\dd \bs{v}}{\dd s} &= f_\psi(\bs{q},\bs{v},s,\bs{x}_t),
\end{aligned}
\label{eq:second_order_dyn}
\end{equation}
where \(\bs{x}_t\) acts as the current physical anchor for action refinement. The central learning problem is to estimate an acceleration field that maps a coarse history-conditioned action prior to a smooth expert-consistent action chunk.

\section{Method}

\subsection{Full-History State Tokenization and Causal State-Space Memory}

The core design of \method\ is to separate three roles that are often entangled in visuomotor policies: perception, historical state estimation, and action generation.
The perception module forms a single state-representative token from the observation and proprioception at each physical timestep. The temporal module propagates a causal recurrent state over the full trajectory. The action module first samples a coarse multimodal prior conditioned on history and current state, and then refines it through second-order dynamics.

Formally, the policy factorizes as
\begin{equation}
\begin{aligned}
    \bs{x}_t &= \phi_\eta(o_t,p_t),\\
    (\bs{h}_t,\bs{y}_t) &= \mathcal{S}_\theta(\bs{h}_{t-1},\bs{x}_t),\\
    \bs{c}_t &= \Gamma_\omega(\bs{y}_t,\bs{x}_t),\\
    \bs{q}_0 &= G_\phi(\bs{z}_t,\bs{c}_t),
    \qquad
    \bs{z}_t\sim\mathcal{N}(\bs{0},\bs{I}),\\
    \hat{\bs{q}}_t &= \mathcal{B}_\psi(\bs{q}_0,\bs{x}_t).
\end{aligned}
\label{eq:policy_factorization}
\end{equation}
Here, \(\bs{x}_t\) is the state token, \(\bs{h}_t\) is the internal recurrent state of the selective SSM, \(\bs{y}_t\) is the emitted causal historical context, \(\bs{c}_t\) is the action-generation condition, \(\bs{q}_0\) is the coarse multimodal action prior, and \(\mathcal{B}_\psi\) denotes the second-order Schr\"odinger-inspired bridge refinement.
The proprioceptive state \(p_t\) is fused inside the token extractor \(\phi_\eta\), rather than passed again as a separate input to the action generator.

The deployment policy is causal: at time \(t\), both \(\bs{h}_t\) and \(\bs{y}_t\) depend only on observations up to \(t\).
During training, complete demonstration sequences are used to preserve temporal gradients through the full recurrent computation graph.

\paragraph*{Modality-agnostic state token extraction}

At each time step, the robot receives an exteroceptive observation \(o_t\) and a proprioceptive state \(p_t\).
The purpose of the token extractor is to produce a compact state-representative embedding for temporal state estimation:
\begin{equation}
    \bs{x}_t=\phi_\eta(o_t,p_t)\in\R^{d}.
    \label{eq:token_general}
\end{equation}
The extractor can be instantiated with either point-cloud or image observations while exposing the same temporal interface: one state token per physical control step.

For point-cloud observations, let
\begin{equation}
    \bs{P}_t=\{\bs{r}_{t,i}\}_{i=1}^{N},
    \qquad
    \bs{r}_{t,i}\in\R^{d_r}.
    \label{eq:point_cloud}
\end{equation}
A shared point encoder maps each point to a local feature,
\begin{equation}
    \bs{u}_{t,i}=\eta_{\mathrm{pc}}(\bs{r}_{t,i}),
    \label{eq:point_local_feat}
\end{equation}
and a permutation-invariant pooling operator forms a global geometric descriptor:
\begin{equation}
    g_t^{\mathrm{pc}}
    =
    \rho_{\mathrm{pc}}
    \left(
        \mathrm{Pool}_{i=1}^{N}\,\bs{u}_{t,i}
    \right).
    \label{eq:pc_global_feat}
\end{equation}
Proprioception is encoded as
\begin{equation}
    g_t^{\mathrm{prop}}=\rho_{\mathrm{prop}}(p_t),
    \label{eq:prop_feat}
\end{equation}
and the fused token is
\begin{equation}
    \bs{x}_t
    =
    \mathrm{LN}
    \left(
    W_x
    \begin{bmatrix}
        g_t^{\mathrm{pc}}\\
        g_t^{\mathrm{prop}}
    \end{bmatrix}
    +b_x
    \right).
    \label{eq:pc_token}
\end{equation}

For image observations, a visual encoder extracts dense features
\begin{equation}
    \bs{F}_t = E_{\mathrm{img}}(o_t)\in\R^{M\times d_f},
    \label{eq:image_features}
\end{equation}
where \(M\) is the number of spatial tokens.
A lightweight adapter compresses these dense features into a global visual descriptor:
\begin{equation}
    g_t^{\mathrm{img}}
    =
    \rho_{\mathrm{img}}
    \left(
        \mathrm{Adapter}(\bs{F}_t)
    \right).
    \label{eq:image_global}
\end{equation}
The image-based state token is
\begin{equation}
    \bs{x}_t
    =
    \mathrm{LN}
    \left(
    W_x
    \begin{bmatrix}
        g_t^{\mathrm{img}}\\
        g_t^{\mathrm{prop}}
    \end{bmatrix}
    +b_x
    \right).
    \label{eq:image_token}
\end{equation}
Thus, different sensing modalities are converted into the same full-history sequence representation.

\begin{figure}[t]
    \centering
    \includegraphics[width=0.98\columnwidth]{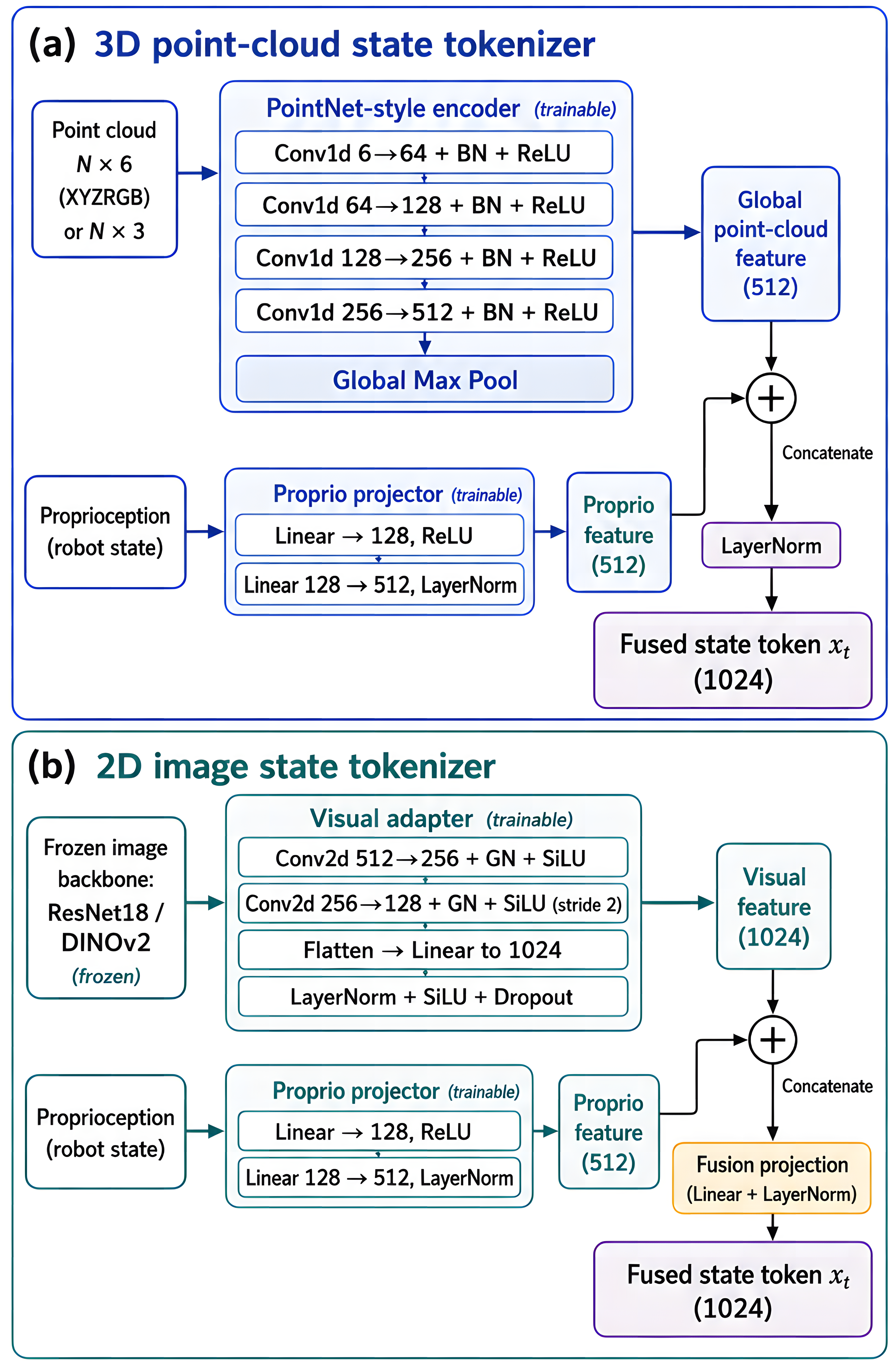}
    \caption{Observation-token extraction branches. Point-cloud and image observations are converted into compact state tokens by modality-specific encoders and a shared proprioceptive fusion interface. Both branches output one state token per physical control step.}
    \label{fig:token_backbones}
\end{figure}

\paragraph*{Causal full-history selective state-space encoding}

The full-history encoder maps the token sequence \(\bs{x}_{1:t}\) into an internal recurrent state \(\bs{h}_t\) and an emitted historical context \(\bs{y}_t\).
We follow the standard state-space convention that separates the recurrent state from the output.
A continuous-time state model has the form
\begin{equation}
\begin{aligned}
    \dot{\bs{h}}(\tau) &= \bs{A}\bs{h}(\tau)+\bs{B}\bs{u}(\tau),\\
    \bs{y}(\tau) &= \bs{C}\bs{h}(\tau)+\bs{D}\bs{u}(\tau),
\end{aligned}
\label{eq:continuous_ssm_revised}
\end{equation}
where \(\bs{h}(\tau)\) is the internal recurrent state, \(\bs{u}(\tau)\) is the input, and \(\bs{y}(\tau)\) is the emitted output.
In a selective state space model, the discretization and input-output projections depend on the current token.
For token \(\bs{x}_t\), we compute
\begin{equation}
    \Delta_t,\bs{B}_t,\bs{C}_t = \Psi_\theta(\bs{x}_t),
    \label{eq:ssm_selective_params}
\end{equation}
and discretize the transition as
\begin{equation}
\begin{aligned}
    \bar{\bs{A}}_t &= \exp(\Delta_t \bs{A}),\\
    \bar{\bs{B}}_t &= \int_{0}^{\Delta_t}
    \exp((\Delta_t-r)\bs{A})\bs{B}_t\,\mathrm{d}r .
\end{aligned}
\label{eq:ssm_discretization_revised}
\end{equation}
The causal recurrence is
\begin{equation}
\begin{aligned}
    \bs{h}_t &= \bar{\bs{A}}_t \bs{h}_{t-1}+\bar{\bs{B}}_t \bs{x}_t,\\
    \bs{y}_t &= \bs{C}_t \bs{h}_t + \bs{D} \bs{x}_t .
\end{aligned}
\label{eq:ssm_recurrence_revised}
\end{equation}
The key property is causal historical dependence:
\begin{equation}
    \bs{y}_t = \mathcal{S}_\theta(\bs{x}_{1:t}),
    \qquad
    \frac{\partial \bs{y}_t}{\partial \bs{x}_\tau}\neq 0
    \quad \text{for some } \tau<t .
    \label{eq:causal_history_context}
\end{equation}
The context \(\bs{y}_t\) summarizes all available past observations and proprioceptive states through \(\bs{h}_t\), while never accessing future tokens at deployment.

During training, losses from all time steps are accumulated over the complete demonstration sequence:
\begin{equation}
    \mathcal{L}(\tau)
    =
    \sum_{t=1}^{L}
    \ell_t(\bs{y}_t,\bs{x}_t,a_{t:t+H-1}^\star),
    \label{eq:trajectory_loss_sum}
\end{equation}
and gradients are backpropagated through the full recurrent computation graph:
\begin{equation}
    \nabla_\theta \mathcal{L}(\tau)
    =
    \sum_{t=1}^{L}
    \frac{\partial \ell_t}{\partial \bs{y}_t}
    \frac{\partial \bs{y}_t}{\partial \theta}.
    \label{eq:full_temporal_gradient}
\end{equation}
This allows late-stage imitation errors to update the earlier recurrent dynamics that formed the latent task phase.
In contrast, detached recurrent training propagates memory forward but cuts part of the temporal credit-assignment path.

\paragraph*{Condition construction and physical anchoring}

The action generator uses accumulated history and current physical evidence in different ways.
The IMLE prior is conditioned on both the emitted historical context \(\bs{y}_t\) and the current state token \(\bs{x}_t\):
\begin{equation}
    \bs{c}_t=\Gamma_\omega(\bs{y}_t,\bs{x}_t),
    \label{eq:condition_revised}
\end{equation}
where \(\Gamma_\omega\) is a lightweight fusion network.
A simple implementation is
\begin{equation}
    \bs{c}_t
    =
    \mathrm{LN}
    \left(
        \bs{W}_c
        \begin{bmatrix}
            \bs{y}_t\\
            \bs{x}_t
        \end{bmatrix}
        +\bs{b}_c
    \right).
    \label{eq:condition_mlp}
\end{equation}
The historical context \(\bs{y}_t\) provides task-phase memory, while \(\bs{x}_t\) preserves the current geometric, visual, and proprioceptive evidence.

The second-order bridge uses \(\bs{x}_t\) as the physical anchoring condition:
\begin{equation}
    \hat{\bs{a}}_s=f_\psi(\bs{q}_s,\bs{v}_s,s,\bs{x}_t).
    \label{eq:anchored_acceleration}
\end{equation}
Thus, the coarse action \(\bs{q}_0\) is generated from the history-conditioned IMLE prior, while the acceleration refinement is anchored to the instantaneous fused state token.

\subsection{Physics-Informed Second-Order Action Bridge}

\paragraph*{Coarse multimodal action prior}

Robotic demonstrations are often multimodal: under the same historical context \(\bs{y}_t\) and current state token \(\bs{x}_t\), several action chunks may be valid because the robot can approach, grasp, or place an object through different feasible motion patterns. Direct regression tends to average these modes, producing actions that may lie between valid behaviors. We therefore use implicit maximum likelihood estimation (IMLE) to learn a conditional implicit generator that represents a multimodal action prior.

Let the expert action chunk at time \(t\) be
\begin{equation}
\begin{aligned}
    q_t^\star
    &=
    \left[
        a_t^\star,
        a_{t+1}^\star,
        \ldots,
        a_{t+H-1}^\star
    \right]\in\R^{D},\\
    D&=H d_a .
\end{aligned}
\label{eq:expert_chunk}
\end{equation}
The IMLE condition is
\begin{equation}
    c_t=\Gamma_\omega(y_t,x_t),
    \label{eq:imle_condition}
\end{equation}
where \(\bs{y}_t\) provides the historical task phase and \(\bs{x}_t\) provides the current fused visual, geometric, and proprioceptive evidence. The IMLE generator maps a latent noise sample and the condition to an action chunk:
\begin{equation}
    \tilde q_t = G_\phi(z_t,c_t),
    \qquad
    z_t \sim p(z)=\mathcal{N}(0,I).
    \label{eq:imle_generator}
\end{equation}
This defines an implicit conditional action distribution:
\begin{equation}
\begin{aligned}
    p_\phi(q\mid c_t)
    =
    \int
    \delta\!\left(q-G_\phi(z,c_t)\right)
    p(z)\,\mathrm{d}z .
\end{aligned}
\label{eq:implicit_pushforward}
\end{equation}
Equivalently, since \(c_t=\Gamma_\omega(y_t,x_t)\), the generator represents \(p_\phi(q\mid y_t,x_t)\). Because \(G_\phi\) is nonlinear, different regions of the latent space can be mapped to different action modes. For a multimodal expert distribution
\begin{equation}
    p_{\mathrm{data}}(q\mid c_t)
    =
    \sum_{j=1}^{J}
    \alpha_j(c_t)\,p_j(q\mid c_t),
    \label{eq:multimodal_expert}
\end{equation}
the generator can allocate disjoint latent regions \(\mathcal{Z}_j(c_t)\) to distinct modes:
\begin{equation}
\begin{aligned}
    G_\phi(z,c_t)
    &\in
    \operatorname{supp}\!\left(p_j(q\mid c_t)\right),\\
    z&\in \mathcal{Z}_j(c_t).
\end{aligned}
\label{eq:latent_mode_partition}
\end{equation}

During training, we draw \(K\) latent samples for each condition:
\begin{equation}
\begin{aligned}
    z_t^{(1)},\ldots,z_t^{(K)}
    &\overset{\mathrm{i.i.d.}}{\sim}\mathcal{N}(0,I),\\
    \tilde q_t^{(k)}
    &=G_\phi(z_t^{(k)},c_t).
\end{aligned}
\label{eq:imle_samples}
\end{equation}
Instead of forcing every generated sample to match the expert chunk, IMLE only requires that at least one generated sample be close to each expert example. The nearest generated candidate is selected by
\begin{equation}
\begin{aligned}
    k^\star
    &=
    \argmin_{1\leq k\leq K}
    d\!\left(\tilde q_t^{(k)},q_t^\star\right),\\
    q_0
    &=\tilde q_t^{(k^\star)},
\end{aligned}
\label{eq:imle_nearest}
\end{equation}
where \(d(\cdot,\cdot)\) is an action-chunk distance, implemented as an \(\ell_1\) or \(\ell_2\) distance. The IMLE objective is
\begin{equation}
\begin{aligned}
    \mathcal{L}_{\mathrm{IMLE}}
    =
    \E_{\substack{(q_t^\star,c_t)\\ z_t^{(1:K)}}}
    \Big[
        \min_{1\leq k\leq K}
        d\!\left(
            G_\phi(z_t^{(k)},c_t),
            q_t^\star
        \right)
    \Big].
\end{aligned}
\label{eq:imle_objective}
\end{equation}
Equivalently, using the nearest sample \(k^\star\),
\begin{equation}
    \mathcal{L}_{\mathrm{IMLE}}
    =
    \E_{\substack{(q_t^\star,c_t)\\ z_t^{(1:K)}}}
    \left[
        d\!\left(
            \tilde q_t^{(k^\star)},
            q_t^\star
        \right)
    \right].
    \label{eq:imle_objective_nearest}
\end{equation}
This nearest-sample objective avoids the mode-averaging behavior of deterministic regression. Each expert action chunk only pulls its closest generated sample, allowing other samples to cover other feasible modes under the same historical and current-state condition. As \(K\) increases, the minimum in \eqref{eq:imle_objective} better approximates the distance from the expert chunk to the generated conditional support, encouraging mode coverage rather than mode collapse.

The selected candidate \(\bs{q}_0\) is not used as the final action. It serves as a coarse multimodal prior for the subsequent second-order bridge. The second-order Schr\"odinger-inspired bridge then refines \(\bs{q}_0\) toward the expert-consistent action chunk by predicting an acceleration field anchored by the current fused state token \(\bs{x}_t\).

\paragraph*{From Schr\"odinger dynamics to quantum potential}

We now derive the second-order bridge objective. Let the action coordinate be \(q\in\R^D\). The Schr\"odinger equation over bridge time \(s\) is
\begin{equation}
    i\hbar\frac{\partial\psi}{\partial s}
    =-\frac{\hbar^2}{2m}\nabla^2\psi+V\psi .
    \label{eq:schrodinger}
\end{equation}
Apply the Madelung transformation
\begin{equation}
\begin{aligned}
    \psi(q,s)
    &=R(q,s)\exp\left(\frac{iS(q,s)}{\hbar}\right),\\
    \rho(q,s)&=R(q,s)^2=|\psi(q,s)|^2 .
\end{aligned}
\label{eq:madelung}
\end{equation}
The required derivatives are
\begin{align}
    \frac{\partial\psi}{\partial s}
    &=\left(
        \frac{\partial R}{\partial s}
        +\frac{i}{\hbar}R\frac{\partial S}{\partial s}
      \right)e^{iS/\hbar},
      \label{eq:mad_dt}\\
    \nabla\psi
    &=\left(\nabla R+\frac{i}{\hbar}R\nabla S\right)e^{iS/\hbar},
      \label{eq:mad_grad}\\
    \nabla^2\psi
    &=\Bigg(
        \nabla^2R
        +\frac{2i}{\hbar}\nabla R\cdot\nabla S
        +\frac{i}{\hbar}R\nabla^2S
        \nonumber\\
    &\qquad
        -\frac{1}{\hbar^2}R\|\nabla S\|^2
      \Bigg)e^{iS/\hbar}.
      \label{eq:mad_lap}
\end{align}
Substituting \eqref{eq:mad_dt}--\eqref{eq:mad_lap} into \eqref{eq:schrodinger} and canceling \(e^{iS/\hbar}\) gives
\begin{equation}
\begin{aligned}
    i\hbar\frac{\partial R}{\partial s}
    -R\frac{\partial S}{\partial s}
    &=-\frac{\hbar^2}{2m}\nabla^2R
      -\frac{i\hbar}{m}\nabla R\cdot\nabla S \\
    &\quad
      -\frac{i\hbar}{2m}R\nabla^2S
      +\frac{1}{2m}R\|\nabla S\|^2
      +VR .
\end{aligned}
\label{eq:substituted}
\end{equation}
Taking imaginary parts yields
\begin{equation}
    \hbar\frac{\partial R}{\partial s}
    =-\frac{\hbar}{2m}
      \left(2\nabla R\cdot\nabla S+R\nabla^2S\right).
    \label{eq:imaginary_part}
\end{equation}
Multiplying by \(2R/\hbar\) and using \(\rho=R^2\), we obtain the continuity equation
\begin{equation}
    \frac{\partial\rho}{\partial s}
    +\nabla\cdot\left(\rho\frac{\nabla S}{m}\right)=0,
    \qquad
    v=\frac{\nabla S}{m}.
    \label{eq:continuity}
\end{equation}
Taking real parts of \eqref{eq:substituted} yields
\begin{equation}
    -R\frac{\partial S}{\partial s}
    =-\frac{\hbar^2}{2m}\nabla^2R
      +\frac{1}{2m}R\|\nabla S\|^2+VR.
    \label{eq:real_part_raw}
\end{equation}
Dividing by \(R\) gives the quantum Hamilton--Jacobi equation
\begin{equation}
    \frac{\partial S}{\partial s}
    +\frac{\|\nabla S\|^2}{2m}
    +V+Q=0,
    \label{eq:quantum_hj}
\end{equation}
where the quantum potential is
\begin{equation}
    Q(q,s)=-\frac{\hbar^2}{2m}\frac{\nabla^2R}{R}.
    \label{eq:quantum_potential}
\end{equation}

\paragraph*{Quantum potential in density form}

Because \(R=\rho^{1/2}\),
\begin{equation}
    \nabla R
    =\nabla(\rho^{1/2})
    =\frac{1}{2}\rho^{-1/2}\nabla\rho .
    \label{eq:grad_R}
\end{equation}
The Laplacian is
\begin{equation}
\begin{aligned}
    \nabla^2R
    &=\nabla\cdot\left(
        \frac{1}{2}\rho^{-1/2}\nabla\rho
      \right)\\
    &=\frac{1}{2}\left[
        \nabla(\rho^{-1/2})\cdot\nabla\rho
        +\rho^{-1/2}\nabla^2\rho
      \right]\\
    &=\frac{1}{2}\rho^{-1/2}\nabla^2\rho
      -\frac{1}{4}\rho^{-3/2}\|\nabla\rho\|^2 .
\end{aligned}
\label{eq:lap_R}
\end{equation}
Therefore
\begin{equation}
    \frac{\nabla^2R}{R}
    =\frac{1}{2}\frac{\nabla^2\rho}{\rho}
     -\frac{1}{4}\frac{\|\nabla\rho\|^2}{\rho^2}.
    \label{eq:lap_R_over_R}
\end{equation}
Using
\begin{equation}
    \nabla\log\rho=\frac{\nabla\rho}{\rho}
    \label{eq:grad_log_rho}
\end{equation}
and
\begin{equation}
    \nabla^2\log\rho
    =\frac{\nabla^2\rho}{\rho}
     -\frac{\|\nabla\rho\|^2}{\rho^2},
    \label{eq:lap_log_rho}
\end{equation}
we have
\begin{equation}
    \frac{\nabla^2\rho}{\rho}
    =\nabla^2\log\rho+\|\nabla\log\rho\|^2.
    \label{eq:rho_identity}
\end{equation}
Substituting \eqref{eq:rho_identity} into \eqref{eq:lap_R_over_R},
\begin{equation}
    \frac{\nabla^2R}{R}
    =\frac{1}{2}\nabla^2\log\rho
     +\frac{1}{4}\|\nabla\log\rho\|^2.
    \label{eq:lap_R_log}
\end{equation}
Thus
\begin{equation}
    Q(q,s)
    =-\frac{\hbar^2}{4m}
      \left(
        \nabla^2\log\rho
        +\frac{1}{2}\|\nabla\log\rho\|^2
      \right).
    \label{eq:Q_log}
\end{equation}
This is the key bridge between probability geometry and acceleration: density gradients and curvature create a force field through \(-\nabla Q\).

\paragraph*{Newtonian projection}

Taking \(\nabla\) of \eqref{eq:quantum_hj} gives
\begin{equation}
    \frac{\partial(\nabla S)}{\partial s}
    +\nabla\left(\frac{\|\nabla S\|^2}{2m}\right)
    =-\nabla V-\nabla Q .
    \label{eq:grad_hj}
\end{equation}
With momentum \(p=\nabla S=mv\), the material derivative form is
\begin{equation}
    \frac{Dp}{Ds}=-\nabla V-\nabla Q.
    \label{eq:momentum_dyn}
\end{equation}
Equivalently,
\begin{equation}
\begin{aligned}
    m a &= F_{\mathrm{classic}}+F_{\mathrm{quantum}},\\
    F_{\mathrm{classic}}&=-\nabla V,\\
    F_{\mathrm{quantum}}&=-\nabla Q .
\end{aligned}
\label{eq:newton_quantum}
\end{equation}
Equation~\eqref{eq:newton_quantum} is the physical reason we predict acceleration instead of only action or velocity.

\paragraph*{Cubic reference bridge}

Let \(\bs{q}_0\) be the IMLE coarse action and \(q_1=q_t^\star\) be the expert action chunk.
The reference bridge follows the Euclidean displacement path between the two action chunks, which is the deterministic optimal-transport (OT) geodesic in action space.
However, the constant-speed OT interpolation \(q_0+s(q_1-q_0)\) is first-order: it carries nonzero bridge-time velocity at the endpoints and is therefore incompatible with a second-order acceleration bridge.
We keep the same OT displacement direction but reparameterize the bridge time with the cubic Hermite clock
\begin{equation}
    q_{\mathrm{ref}}(s)
    =
    q_0+\alpha(s)(q_1-q_0),
    \quad
    \alpha(s)=3s^2-2s^3 .
    \label{eq:cubic_ref}
\end{equation}
This is the lowest-degree monotone time reparameterization satisfying
\[
    \alpha(0)=0,\quad
    \alpha(1)=1,\quad
    \dot\alpha(0)=0,\quad
    \dot\alpha(1)=0,
\]
so it preserves the OT displacement path while enforcing the position and velocity boundary conditions required by second-order dynamics.
Its velocity and acceleration are
\begin{equation}
\begin{aligned}
    v_{\mathrm{ref}}(s)
    &=6s(1-s)(q_1-q_0),\\
    a_{\mathrm{spline}}(s)
    &=(6-12s)(q_1-q_0).
\end{aligned}
\label{eq:cubic_vel_acc}
\end{equation}
We interpret the classical force as a time-moving potential well whose nominal acceleration is
\begin{equation}
    F_{\mathrm{classic}}
    =
    -\nabla V_{\mathrm{ext}}
    =
    m\,a_{\mathrm{spline}}(s).
    \label{eq:classic_force}
\end{equation}

\paragraph*{Gaussian quantum control projection}

Around the cubic path, assume a local Gaussian density
\begin{equation}
    \rho(q,s)
    =\frac{1}{(2\pi\sigma_s^2)^{D/2}}
      \exp\left(
        -\frac{\|q-q_{\mathrm{ref}}(s)\|^2}{2\sigma_s^2}
      \right).
    \label{eq:gaussian_density}
\end{equation}
Then
\begin{equation}
    R(q,s)\propto
      \exp\left(
        -\frac{\|q-q_{\mathrm{ref}}(s)\|^2}{4\sigma_s^2}
      \right).
    \label{eq:gaussian_R}
\end{equation}
The first derivative is
\begin{equation}
    \nabla R
    =-\frac{q-q_{\mathrm{ref}}}{2\sigma_s^2}R.
    \label{eq:gaussian_grad_R}
\end{equation}
The Laplacian is
\begin{equation}
    \nabla^2R
    =\left(
        \frac{\|q-q_{\mathrm{ref}}\|^2}{4\sigma_s^4}
        -\frac{D}{2\sigma_s^2}
      \right)R.
    \label{eq:gaussian_lap_R}
\end{equation}
Substituting into \eqref{eq:quantum_potential},
\begin{equation}
    Q(q,s)
    =-\frac{\hbar^2}{2m}
      \left(
        \frac{\|q-q_{\mathrm{ref}}\|^2}{4\sigma_s^4}
        -\frac{D}{2\sigma_s^2}
      \right).
    \label{eq:gaussian_Q}
\end{equation}
The free-particle Gaussian quantum term is dispersive. For robotic control we use its quadratic structure as a stabilizing projection around the reference path, absorbing constants and sign into a learned or chosen stiffness:
\begin{equation}
    F_{q}(q,s)=K_p\big(q_{\mathrm{ref}}(s)-q\big).
    \label{eq:pd_quantum_position}
\end{equation}
This is the position-stabilizing component of the acceleration target.

\paragraph*{Kostin dissipation and velocity stabilization}

The conservative Schr\"odinger equation can oscillate indefinitely around a potential well. Kostin introduced a dissipative potential
\begin{equation}
    V_{\mathrm{diss}}
    =\frac{\hbar\gamma}{2i}
      \ln\left(\frac{\psi}{\psi^\ast}\right).
    \label{eq:kostin_potential}
\end{equation}
Using \eqref{eq:madelung},
\begin{equation}
\begin{aligned}
    \ln\left(\frac{\psi}{\psi^\ast}\right)
    &=\ln\left(
        \frac{R e^{iS/\hbar}}{R e^{-iS/\hbar}}
      \right)\\
    &=\ln\left(e^{2iS/\hbar}\right)
      =\frac{2iS}{\hbar}.
\end{aligned}
\label{eq:kostin_log}
\end{equation}
Hence
\begin{equation}
    V_{\mathrm{diss}}=\gamma S.
    \label{eq:kostin_gammaS}
\end{equation}
The Hamilton--Jacobi equation becomes
\begin{equation}
    \frac{\partial S}{\partial s}
    +\frac{\|\nabla S\|^2}{2m}
    +V_{\mathrm{classic}}+Q+\gamma S=0.
    \label{eq:kostin_hj}
\end{equation}
Taking the Newtonian projection,
\begin{equation}
    m a
    =-\nabla V_{\mathrm{classic}}
     -\nabla Q
     -\gamma\nabla S.
    \label{eq:kostin_newton}
\end{equation}
Since \(\nabla S=mv\), the last term is viscous friction. Centering it around the reference velocity gives
\begin{equation}
    F_v(v,s)=K_d\big(v_{\mathrm{ref}}(s)-v\big).
    \label{eq:velocity_force}
\end{equation}
The resulting target acceleration is
\begin{equation}
    a_{\mathrm{tar}}(q,v,s)
    =a_{\mathrm{spline}}(s)
     +K_p\big(q_{\mathrm{ref}}(s)-q\big)
     +K_d\big(v_{\mathrm{ref}}(s)-v\big).
    \label{eq:target_acc}
\end{equation}
This expression has the form of a generalized second-order stabilizer around a cubic reference path. The role of the Schr\"odinger--Madelung derivation is not to claim an exact solution of a generic bridge problem, but to motivate an acceleration-level control projection with position stabilization and dissipative velocity correction.

\paragraph*{Second-order noise schedule}

A noisy bridge state is sampled as
\begin{equation}
\begin{aligned}
    q_s
    &=
    q_{\mathrm{ref}}(s)+\sigma(s)\epsilon,\\
    v_s
    &=
    v_{\mathrm{ref}}(s)+\dot\sigma(s)\epsilon,
\end{aligned}
\qquad
\epsilon\sim\mathcal{N}(0,I).
\label{eq:noisy_bridge}
\end{equation}
The key difference from first-order denoising is that the same perturbation affects both the position-level action chunk \(q_s\) and the bridge velocity \(v_s\).
Therefore, the noise schedule must satisfy second-order boundary conditions:
\begin{equation}
    \sigma(0)=\sigma(1)=0,
    \qquad
    \dot\sigma(0)=\dot\sigma(1)=0 .
    \label{eq:noise_boundary}
\end{equation}
The first pair of constraints ensures that the noisy position remains anchored to the IMLE prior at \(s=0\) and to the expert chunk at \(s=1\).
The second pair ensures that the noisy bridge does not introduce random endpoint velocities.
These velocity constraints are essential because the network is trained to predict acceleration.
If the endpoint velocity noise is nonzero, the acceleration target must implicitly compensate for an artificial boundary impulse rather than only learning smooth action refinement.

A first-order diffusion-style bridge schedule does not satisfy these requirements.
For example, the Brownian bridge schedule
\begin{equation}
    \sigma_{\mathrm{BB}}(s)
    \propto
    \sqrt{s(1-s)}
\end{equation}
vanishes at the two endpoints, but its derivative is singular as \(s\to0\) or \(s\to1\).
The simple parabolic schedule
\begin{equation}
    \sigma_{\mathrm{par}}(s)
    =
    \sigma_{\max}s(1-s)
\end{equation}
has finite but nonzero endpoint derivatives:
\begin{equation}
    \dot\sigma_{\mathrm{par}}(0)=\sigma_{\max},
    \qquad
    \dot\sigma_{\mathrm{par}}(1)=-\sigma_{\max}.
\end{equation}
Thus, although such schedules are natural for first-order score or denoising models, they are mismatched to the present second-order bridge because they inject random velocity at the bridge boundaries.
In practice, this can produce boundary jitter, unstable contact approach, and inaccurate insertion behavior: the learned acceleration field must spend capacity damping scheduler-induced endpoint velocity rather than refining the coarse action prior.

We therefore use the quartic bell schedule
\begin{equation}
    \sigma(s)
    =
    16\sigma_{\max}s^2(1-s)^2,
    \label{eq:quartic_sigma}
\end{equation}
whose derivative is
\begin{equation}
    \dot\sigma(s)
    =
    32\sigma_{\max}s(1-s)(1-2s).
    \label{eq:quartic_sigma_dot}
\end{equation}
This schedule satisfies
\begin{equation}
    \sigma(0)=\sigma(1)=0,
    \qquad
    \dot\sigma(0)=\dot\sigma(1)=0,
\end{equation}
and reaches its maximum at the middle of the bridge.
It is the lowest-degree polynomial bell-shaped schedule that simultaneously anchors position and velocity at both endpoints.
This makes the training perturbation consistent with the second-order boundary-value structure of the acceleration bridge.

\subsection{Memory-Efficient Training and Causal Inference}

\paragraph*{Memory-efficient perception--then--SSM training}
The main memory bottleneck in long-horizon visuomotor learning is usually high-dimensional perception rather than the recurrent state-space computation. A naive image-based implementation that backpropagates through a high-capacity visual backbone over all \(L\) time steps must retain visual activations for every frame:
\begin{equation}
\begin{aligned}
    \mathcal{M}_{\mathrm{naive}}
    \approx\;&
    \mathcal{O}\!\left(L\,\mathcal{M}_{\mathrm{bb,bwd}}\right)\\
    &+
    \mathcal{O}\!\left(L\,\mathcal{M}_{\mathrm{temp}}\right),
\end{aligned}
\label{eq:naive_memory}
\end{equation}
where \(\mathcal{M}_{\mathrm{bb,bwd}}\) denotes the activation memory required to train the high-dimensional perceptual backbone, and \(\mathcal{M}_{\mathrm{temp}}\) denotes the memory of the trainable SSM and action-generation modules.

\method\ separates the high-dimensional perception cost from full-history temporal learning. Given a perception chunk size \(C\), the trajectory is divided into chunks
\begin{equation}
    \mathcal{O}_b
    =
    \{(o_t,p_t)\}_{t=bC+1}^{\min((b+1)C,L)} .
    \label{eq:perception_chunk}
\end{equation}
Each chunk is encoded into low-dimensional state tokens:
\begin{equation}
    \bs{X}_b
    =
    \phi_\eta(\mathcal{O}_b)
    =
    \{\bs{x}_{bC+1},\ldots,\bs{x}_{\min((b+1)C,L)}\}.
    \label{eq:chunk_token_extract}
\end{equation}
The full token sequence is then assembled and passed through the temporal SSM once:
\begin{equation}
\begin{aligned}
    \bs{X}_{1:L}
    &=
    \mathrm{Concat}(\bs{X}_0,\bs{X}_1,\ldots,\bs{X}_{B-1}),\\
    (\bs{h}_{1:L},\bs{y}_{1:L})
    &=
    \mathcal{S}_\theta(\bs{X}_{1:L}) .
\end{aligned}
\label{eq:full_ssm_after_chunks}
\end{equation}

The memory behavior depends on the perception branch.
In the image-based ALOHA and real-world experiments, the DINOv2 or ResNet18 backbone is frozen and evaluated in no-gradient chunks, so full-trajectory visual-backbone activations are not stored.
Gradients are still preserved through the trainable adapters, fusion layers, SSM, coarse-prior generator, and second-order bridge after the token sequence is assembled.
A representative peak-memory model for this frozen-image setting is
\begin{equation}
\begin{aligned}
    \mathcal{M}_{\mathrm{frz}}
    \approx\;&
    \mathcal{O}\!\left(C\,\mathcal{M}_{\mathrm{bb,fwd}}\right)
    +
    \mathcal{O}\!\left(L\,\mathcal{M}_{\mathrm{adapt}}\right)\\
    &+
    \mathcal{O}\!\left(L\,d\right)
    +
    \mathcal{O}\!\left(L\,\mathcal{M}_{\mathrm{temp}}\right).
\end{aligned}
\label{eq:chunked_memory_revised}
\end{equation}
Here, \(\mathcal{M}_{\mathrm{bb,fwd}}\) is the transient forward memory of the frozen backbone, \(\mathcal{M}_{\mathrm{adapt}}\) is the activation memory of the lightweight adapter, \(d\) is the token dimension, and \(\mathcal{M}_{\mathrm{temp}}\) is the memory of the temporal and action modules.
By removing the dominant high-dimensional backbone cost from full-trajectory backpropagation, this decomposition reduces memory and computation enough to enable full-trajectory parallel training on commodity GPUs.

In the point-cloud RoboTwin 2.0 and RMBench experiments, the PointNet-style encoder is trainable and memory efficient because point clouds are compact geometric inputs.
The same chunked pipeline preserves gradients through the trainable point encoder, temporal SSM, and action-generation modules; for extremely long trajectories, we enable gradient checkpointing across perception chunks.

Importantly, state tokens are not detached before the temporal SSM or action-generation losses. The SSM receives the complete token sequence, and the objective backpropagates through the full recurrent computation graph of the trainable temporal-action stack.Thus, late-phase imitation errors can update the earlier recurrent dynamics that formed the latent task phase, unlike detached recurrent training\cite{mtil}.

\begin{algorithm}[t]
\caption{Chunked Perception--Then--SSM Encoding}
\label{alg:chunked_encoding}
\begin{algorithmic}[1]
\Require Trajectory \(\tau=\{(o_t,p_t,a_t^\star)\}_{t=1}^{L}\), chunk size \(C\)
\State Initialize an empty token list \(\bs{X}\)
\For{\(b=0,\ldots,\lceil L/C\rceil-1\)}
    \State Form perception chunk \(\mathcal{O}_b\)
    \State Encode \(\mathcal{O}_b\) into state tokens \(\bs{X}_b=\phi_\eta(\mathcal{O}_b)\)
    \State Append \(\bs{X}_b\) to \(\bs{X}\)
\EndFor
\State Run the selective SSM over the full token sequence:
\[
    (\bs{h}_{1:L},\bs{y}_{1:L})=\mathcal{S}_\theta(\bs{X}_{1:L})
\]
\State Construct action-generation conditions \(\bs{c}_t=\Gamma_\omega(\bs{y}_t,\bs{x}_t)\) for all \(t\)
\State \Return \(\bs{c}_{1:L}\)
\end{algorithmic}
\end{algorithm}

\paragraph*{Training objective}

\begin{algorithm}[t]
\caption{Training \method}
\label{alg:training}
\begin{algorithmic}[1]
\Require Expert trajectories \(\{\tau\}\), perception chunk size \(C\), IMLE samples \(K\)
\For{each trajectory \(\tau=\{(o_t,p_t,a_t^\star)\}_{t=1}^{L}\)}
    \State Encode observations and proprioception in chunks to obtain \(x_{1:L}\)
    \State Run full-history SSM to obtain internal states and historical contexts:
    \[
        (h_{1:L},y_{1:L})=\mathcal{S}_\theta(X_{1:L})
    \]
    \For{each time step \(t\)}
        \State Form IMLE condition \(c_t=\Gamma_\omega(y_t,x_t)\)
        \State Sample \(z_t^{(1)},\ldots,z_t^{(K)}\sim\mathcal{N}(0,I)\)
        \State Generate candidates \(\tilde q_t^{(k)}=G_\phi(z_t^{(k)},c_t)\)
        \State Select \(\bs{q}_0\) by nearest expert matching
        \State Sample \(s\sim\mathcal{U}(0,1)\), \(\epsilon\sim\mathcal{N}(0,I)\)
        \State Build \(q_{\mathrm{ref}},v_{\mathrm{ref}},a_{\mathrm{spline}}\)
        \State Build \(q_s,v_s\) with the quartic schedule
        \State Compute \(a_{\mathrm{tar}}\) by \eqref{eq:target_acc}
        \State Predict \(\hat a_s=f_\psi(q_s,v_s,s,x_t)\)
    \EndFor
    \State Update all trainable parameters with full temporal backpropagation
\EndFor
\end{algorithmic}
\end{algorithm}

The second-order bridge head predicts an acceleration field anchored by the current fused state token:
\begin{equation}
    \hat a_s=f_\psi(q_s,v_s,s,x_t).
    \label{eq:pred_acc}
\end{equation}
The acceleration matching loss is
\begin{equation}
    \mathcal{L}_{\mathrm{acc}}
    =
    \E_{t,s,\epsilon}
    \left[
        \left\|
          f_\psi(q_s,v_s,s,x_t)
          -
          a_{\mathrm{tar}}(q_s,v_s,s)
        \right\|_2^2
    \right].
    \label{eq:acc_loss}
\end{equation}
The full objective is
\begin{equation}
    \mathcal{L}
    =
    \mathcal{L}_{\mathrm{IMLE}}
    +
    \lambda_{\mathrm{acc}}\mathcal{L}_{\mathrm{acc}}
    +
    \lambda_{\mathrm{bc}}\mathcal{L}_{\mathrm{BC}},
    \label{eq:total_loss}
\end{equation}
with
\begin{equation}
    \mathcal{L}_{\mathrm{BC}}
    =
    \E_t\left[\|\hat q_t-q_t^\star\|_1\right].
    \label{eq:bc_loss}
\end{equation}

\paragraph*{Temporally correlated latent sampling at inference}
Directly resampling the IMLE latent variable independently at every control step can introduce discontinuities in the coarse action prior. During closed-loop deployment, we therefore maintain a temporally correlated latent state \((z_t,\dot z_t)\). At the beginning of each episode, both variables are initialized from a standard Gaussian distribution. At each control step, we update them as
\begin{equation}
\begin{aligned}
    \dot z_t &= \alpha \dot z_{t-1}
    + \sqrt{1-\alpha^2}\,\epsilon_t,
    \qquad \epsilon_t\sim\mathcal{N}(0,I),\\
    \bar z_t &= \cos(\theta) z_{t-1}+\sin(\theta)\dot z_t,\\
    z_t &= \mathrm{Norm}(\bar z_t),
\end{aligned}
\label{eq:latent_ar2}
\end{equation}
where \(\alpha\) controls latent velocity persistence, \(\theta\) controls latent evolution speed, and \(\mathrm{Norm}(\cdot)\) normalizes the latent tensor to zero mean and unit standard deviation within each sample. This update preserves a Gaussian-like latent manifold while making consecutive IMLE priors temporally coherent. The resulting coarse prior is
\begin{equation}
    q_0 = G_\phi(z_t,\Gamma_\omega(y_t,x_t)).
    \label{eq:inference_q0_ar2}
\end{equation}
The prior is then refined by the second-order Schr\"odinger-inspired bridge.

\begin{algorithm}[t]
\caption{Causal Inference with Full-History Memory and Second-Order Refinement}
\label{alg:inference}
\begin{algorithmic}[1]
\Require Initial SSM state \(\bs{h}_0\), IMLE latent state \((\bs{z}_0,\dot{\bs{z}}_0)\), integration steps \(N\)
\For{each control step \(t=1,2,\ldots\)}
    \State Encode current observation and proprioception:
    \[
        x_t=\phi_\eta(o_t,p_t)
    \]
    \State Update the selective SSM:
    \[
        (h_t,y_t)=\mathcal{S}_\theta.\mathrm{step}(h_{t-1},x_t)
    \]
    \State Update temporally correlated IMLE latent variables using \eqref{eq:latent_ar2}
    \State Generate IMLE coarse action prior:
    \[
        q^{(0)}=G_\phi(z_t,\Gamma_\omega(y_t,x_t)),\qquad v^{(0)}=0
    \]
    \For{\(r=0,\ldots,N-1\)}
        \State Predict acceleration:
        \[
            a^{(r)}=f_\psi(q^{(r)},v^{(r)},s_r,x_t)
        \]
        \State Symplectic Euler update:
        \[
        \begin{aligned}
            v^{(r+1)} &= v^{(r)}+\Delta s\,a^{(r)},\\
            q^{(r+1)} &= q^{(r)}+\Delta s\,v^{(r+1)}.
        \end{aligned}
        \]
    \EndFor
    \State Execute the first action or temporally aggregated action from \(\hat q_t=q^{(N)}\)
\EndFor
\end{algorithmic}
\end{algorithm}

At inference, the IMLE generator first samples a history-conditioned coarse prior:
\begin{equation}
    q^{(0)}=G_\phi(z_t,\Gamma_\omega(y_t,x_t)),
    \qquad
    v^{(0)}=0.
    \label{eq:inference_init}
\end{equation}
We then integrate the acceleration field for \(N\) steps. A semi-implicit Euler update for the second-order bridge is
\begin{equation}
\begin{aligned}
    v^{(r+1)}&=v^{(r)}+\Delta s\,
      f_\psi(q^{(r)},v^{(r)},s_r,x_t),\\
    q^{(r+1)}&=q^{(r)}+\Delta s\,v^{(r+1)}.
\end{aligned}
\label{eq:symplectic_euler}
\end{equation}
The final action chunk is \(\hat q=q^{(N)}\). Because the model predicts acceleration, local changes in \(f_\psi\) integrate through velocity before changing position, reducing direct action jitter.

\begin{figure*}[t]
    \centering
    \includegraphics[width=0.96\textwidth]{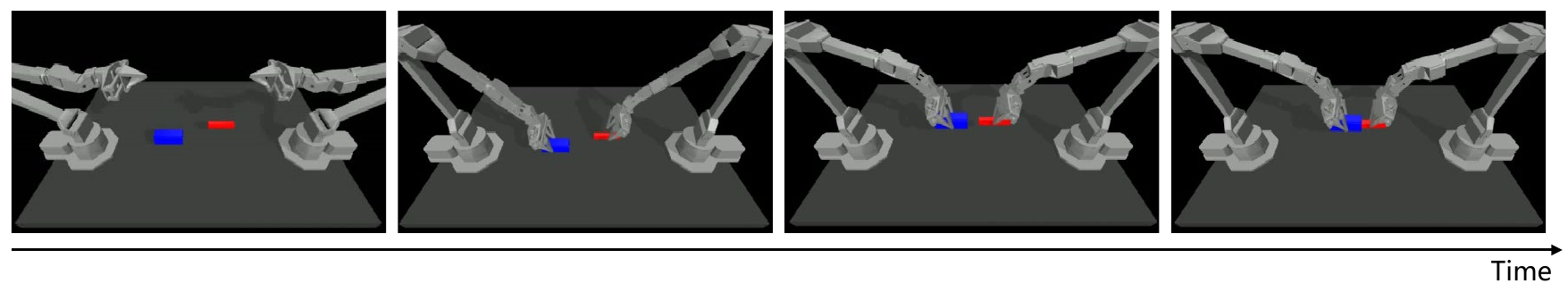}
    \vspace{-1mm}
    \caption{
    Qualitative rollout on the ALOHA bimanual insertion benchmark.
    From left to right, initial scene, pick up blocks, pre-insertion alignment, insertion, and successful seating.
    The full-history encoder and second-order action refinement produce stable bimanual coordination during the precision-critical insertion stage.
    }
    \label{fig:aloha_insertion}
    \vspace{-2mm}
\end{figure*}

\section{Experiments}

\subsection{Benchmarks and Protocol}
We evaluate \method\ on 16 simulated tasks and 4 real-world experiments. The simulated suite consists of one ALOHA bimanual insertion task for isolating fine-grained action generation, eight RoboTwin 2.0 tasks for general manipulation under point-cloud input, and seven RMBench tasks for memory-dependent manipulation. The real-world experiments evaluate whether the learned full-history policy and second-order action refinement transfer to physical robot execution.

The experiments are intended to separate three capabilities that are often conflated: fine-grained action generation, geometric present-state competence, and historical phase competence. The ALOHA bimanual insertion task stresses precise dual-arm coordination and millimeter-level insertion accuracy. RoboTwin 2.0 mostly stresses present-state manipulation under object and scene variation; our model uses point-cloud input on this benchmark, but the benchmark should be interpreted as general manipulation rather than memory-dependent manipulation. RMBench stresses non-Markovian memory, where the correct action depends on historical phase rather than only on the current observation.

\paragraph*{Observation encoders}
We use different perception instantiations according to the benchmark. ALOHA uses a frozen DINOv2 image backbone with trainable adapters. The real-world experiments use a frozen ResNet18 backbone from a single RGB camera view, rather than point clouds, because the D435 point cloud quality is unreliable around small covers and tabletop contact regions. RoboTwin 2.0 and RMBench use 3D point-cloud observations with a trainable PointNet-style encoder. These choices affect perception capacity and should be considered when comparing simulation and real-world results.

\paragraph*{Result sources}
Due to the computational cost of training large VLA and diffusion-based robot policies, we use officially reported baseline results when available. We distinguish three result sources: \emph{Ours}, \emph{Reproduced}, and \emph{Reported}. Results marked as \emph{Reported} are taken from the corresponding papers, benchmark releases, or official model reports. Results marked as \emph{Ours} are obtained by evaluating \method\ under the task definitions and metrics used in the corresponding benchmark. We do not claim that all baselines were retrained under identical random seeds, hardware, observation encoders, or codebases. Therefore, comparisons against reported VLA systems should be interpreted as benchmark-level context, while the Chronos variants and reproduced compact policies provide the most controlled comparisons.

\paragraph*{Metrics}
All reported task numbers are success rates unless otherwise stated. For a task \(i\) with \(N_i\) evaluation rollouts, the success rate is
\begin{equation}
    s_i=\frac{1}{N_i}\sum_{n=1}^{N_i}\mathbb{I}[\mathrm{success}_{i,n}].
    \label{eq:success_rate}
\end{equation}
For a task suite \(\mathcal{T}\), we report the unweighted macro-average
\begin{equation}
    \mathrm{Avg.}=\frac{1}{|\mathcal{T}|}\sum_{i\in\mathcal{T}}s_i .
    \label{eq:macro_average}
\end{equation}
The main comparison tables report point-estimate success rates for all methods because official baseline sources do not consistently provide seed-level statistics. For reported baselines, standard deviations are omitted when they are not provided by the original source; we do not synthesize missing variance estimates for prior work.

\paragraph*{Model scale}
For model scale, we use publicly reported parameter counts whenever available. For ACT, DP, DP3, FlowPolicy, RDT-1B, and \(\pi_0\), we follow reported model sizes from a recent RoboTwin-style comparison. For \(\pi_{0.5}\), X-VLA, and Mem-0, we use their public model-scale descriptions. These counts are used only to contextualize scale, because compact non-VLA policies can vary with encoder, horizon, and implementation. We do not claim that \method\ is smaller than every compact imitation baseline. Instead, \method\ targets a mid-scale regime: comparable to strong 3D generative policies such as DP3, while substantially smaller than VLA-scale and memory-VLA baselines.

\begin{table}[t]
\centering
\caption{Model scale context. ACT, DP, DP3, FlowPolicy, RDT-1B, and \(\pi_0\) follow reported model sizes from a recent RoboTwin-style comparison. Other large-model counts follow public model-scale descriptions.}
\label{tab:param_budget}
\small
\setlength{\tabcolsep}{3.5pt}
\begin{tabularx}{\columnwidth}{lXc}
\toprule
Method & Policy family & Scale \\
\midrule
ACT & Compact action-chunk policy & 80.0M \\
DP & Compact diffusion policy & 96.8M \\
DP3 & 3D diffusion policy & 264.4M \\
RDT-1B & Large robot diffusion transformer & 1.2B \\
\(\pi_0\) & VLA / flow policy & 3.3B \\
\(\pi_{0.5}\) & VLA / flow policy & \(>3.3\)B \\
X-VLA & Cross-embodiment VLA & 0.9B \\
Mem-0 & Memory VLA & \(>10\)B \\
\method & Full-history physics policy & 0.3B \\
\bottomrule
\end{tabularx}
\end{table}

\subsection{Fine-Grained Bimanual Insertion on ALOHA}

We first evaluate action generation on the ALOHA bimanual insertion task, a simulated dual-arm precision manipulation benchmark introduced with the ALOHA system. In this task, the left and right arms grasp a socket and a peg, bring them into contact, and complete insertion in mid-air. The insertion clearance is approximately \(5\,\mathrm{mm}\), making the task sensitive to small action errors and motion jitter. This setting is therefore well suited for isolating whether the action generator can produce precise and smooth action chunks. In this experiment, \method\ uses frozen DINOv2 image features followed by trainable adapters and the full-history temporal/action stack. Figure~\ref{fig:aloha_insertion} visualizes a representative successful rollout.

\begin{table}[t]
    \centering
    \caption{ALOHA bimanual insertion benchmark with 50 demonstrations and 50 evaluation trials. ACT and MTIL are reproduced; Chronos variants share the same history encoder.}
    \label{tab:aloha_ablation}
    \small
    \setlength{\tabcolsep}{3.2pt}
    \renewcommand{\arraystretch}{1.05}
    \begin{tabularx}{\columnwidth}{@{}lXc@{}}
        \toprule
        Method & Action head / schedule & Success \\
        \midrule
        ACT~\cite{act}
            & Regression
            & 50\% \\
        MTIL~\cite{mtil}
            & Regression
            & 76\% \\
        \midrule
        \method\ w/o IMLE/SB
            & Regression
            & 76\% \\
        \method\ w/ diffusion
            & Diffusion
            & 66\% \\
        \method\ w/ flow
            & Flow matching
            & 72\% \\
        \method\ w/o SB
            & IMLE
            & 86\% \\
        \method\ w/o IMLE
            & Regression prior + SB
            & 84\% \\
        \method\ w/ first-order schedule
            & IMLE prior + SB, diffusion scheduler
            & 72\% \\
        \method
            & IMLE prior + SB, quartic scheduler
            & \textbf{90\%} \\
        \bottomrule
        \multicolumn{3}{@{}p{\columnwidth}@{}}{
            \scriptsize
            \emph{Note.}
            SB denotes the proposed second-order Schr\"odinger-inspired bridge.
            The first-order schedule uses a conventional diffusion-style noise schedule that does not enforce zero endpoint velocity in bridge time, whereas the full model uses the quartic schedule in \eqref{eq:quartic_sigma}.
        } \\
    \end{tabularx}
    \vspace{-5mm}
\end{table}

\paragraph*{Interpretation of the action-generation ablation}

The comparison mainly isolates the action head because all \method\ variants share the same history encoder.
The reproduced MTIL baseline also provides a useful reference for detached recurrent training: MTIL reaches 76\%, matching \method\ w/o IMLE/SB under a regression head.
This suggests that, on this precision insertion task, the dominant bottleneck is not long-range memory alone, but whether the action head can convert the historical state into smooth and accurate bimanual motion.

Replacing the second-order bridge with diffusion or flow matching reduces success to 66\% and 72\%, respectively, while removing only the bridge reduces success from 90\% to 86\%.
The schedule ablation further isolates the importance of matching the noise process to second-order dynamics: using the same IMLE-prior second-order bridge but replacing the quartic bell schedule with a conventional first-order diffusion scheduler reduces success from 90\% to 72\%.
This drop indicates that the improvement is not only due to using an acceleration network, but also to using a bridge perturbation that satisfies the position and velocity boundary conditions required by the second-order formulation.

This result illustrates why robot action generation should not directly inherit first-order generative mechanisms from image synthesis.
Diffusion policies perturb action coordinates with stochastic noise and recover them through iterative denoising, which is effective for multimodal generation but can introduce random action-level fluctuations in contact-rich control~\cite{ddpm,scorebased,diffusionpolicy}. Flow matching improves over stochastic denoising by transporting samples with a deterministic first-order velocity field, but the field still updates the action coordinate directly and does not explicitly constrain acceleration or endpoint velocity~\cite{flowmatching,rectifiedflow,pi0}. In contrast, \method\ predicts an acceleration field under a quartic schedule whose position and velocity perturbations vanish at both bridge endpoints. Local corrections therefore affect the final action only after integration through velocity, providing a physically grounded smoothing mechanism. For bimanual insertion, where success depends on a stable approach trajectory rather than only the final pose, this second-order formulation is advantageous.

\begin{table}[t]
    \centering
    \caption{Ablation of inference integration steps on ALOHA bimanual insertion. All variants use the same trained model and differ only in the number of second-order bridge integration steps at inference.}
    \label{tab:aloha_steps}
    \small
    \setlength{\tabcolsep}{3.0pt}
    \renewcommand{\arraystretch}{1.05}
    \begin{tabular*}{\columnwidth}{@{\extracolsep{\fill}}lcccccccc@{}}
        \toprule
        Steps \(N\) & 1 & 2 & 3 & 4 & 5 & 7 & 10 & 15 \\
        \midrule
        Success & 84\% & 82\% & \textbf{90\%} & 84\% & \textbf{90\%} & 86\% & 84\% & 88\% \\
        \bottomrule
    \end{tabular*}
    \vspace{-2mm}
\end{table}

\paragraph*{Effect of inference integration steps}

Table~\ref{tab:aloha_steps} shows that \method\ achieves strong performance with only a few bridge integration steps. A single step already reaches 84\%, and three to five steps achieve the best success rate of 90\%. This indicates that the learned acceleration field acts as a lightweight second-order refinement rather than a long iterative sampler. Compared with diffusion and flow-matching action heads, which typically require around ten or more denoising or velocity-integration steps to generate an action chunk~\cite{diffusionpolicy,pi0,pi05}, \method\ attains peak performance with only 3--5 integration steps.The advantage comes from the structure of the bridge: IMLE first provides a coarse multimodal action prior, while the acceleration field performs local, dynamics-aware refinement around this prior. Thus, computation is concentrated on correcting a plausible action mode instead of repeatedly transporting pure noise to an action trajectory. Increasing the number of steps beyond five brings no additional gain, confirming that the proposed bridge is most effective as a compact, low-latency refinement module for precise robot control.

\begin{figure*}[t]
    \centering
    \includegraphics[width=1\textwidth]{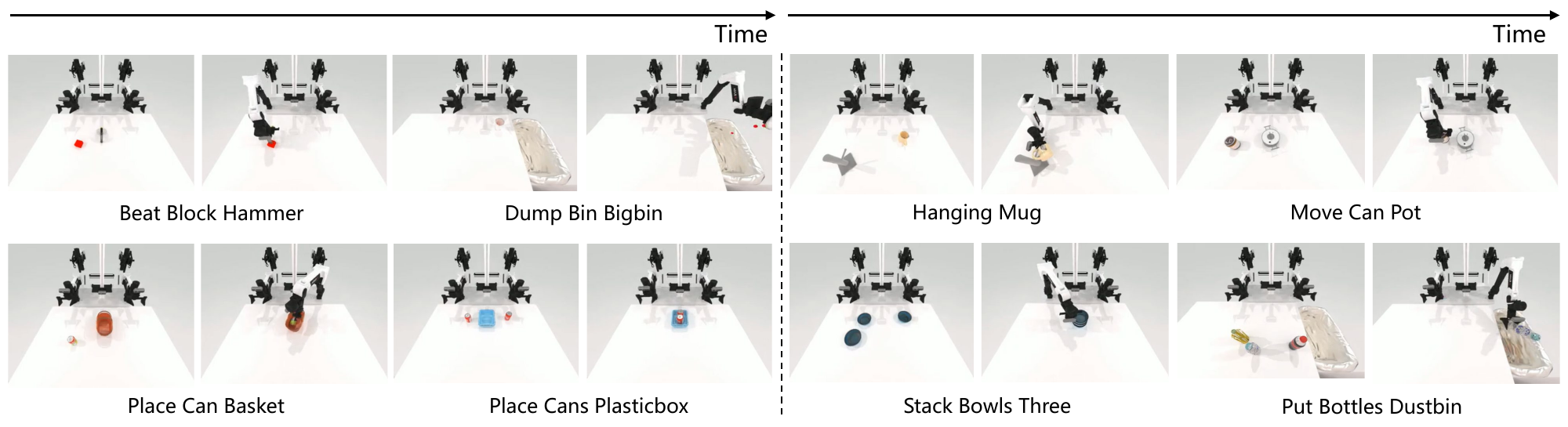}
    \vspace{-1mm}
    \caption{
    Qualitative rollouts on RoboTwin 2.0.
    The eight representative tasks cover contact-rich manipulation, object transfer, container placement, hanging, stacking, and tool-use behaviors.
    Chronos produces coherent action sequences across diverse 3D manipulation scenarios, complementing the quantitative results in Table~\ref{tab:robotwin}.
    }
    \label{fig:robotwin_qualitative}
    \vspace{-2mm}
\end{figure*}

\begin{table*}[t]
    \centering
    \caption{RoboTwin 2.0 Easy-protocol success rates with 50 demonstrations and 100 evaluation trials. Baselines are officially reported results unless otherwise specified.}
    \label{tab:robotwin}
    \small
    \setlength{\tabcolsep}{2.0pt}
    \renewcommand{\arraystretch}{1.05}
    \begin{tabularx}{\textwidth}{@{}lYYYYYY@{}}
        \toprule
        Task
            & \makecell{DP\\(96.8M)}
            & \makecell{ACT\\(80.0M)}
            & \makecell{RDT-1B\\(1.2B)}
            & \makecell{\(\pi_0\)\\(3.3B)}
            & \makecell{DP3\\(264.4M)}
            & \makecell{\method\\(0.3B)} \\
        \midrule
        Hanging Mug
            & 8\% & 7\% & 23\% & 11\% & 17\% & \textbf{26\%} \\
        Place Cans Plasticbox
            & 40\% & 16\% & 6\% & 34\% & 48\% & \textbf{95\%} \\
        Stack Bowls Three
            & 63\% & 48\% & 51\% & 66\% & 57\% & \textbf{67\%} \\
        Put Bottles Dustbin
            & 22\% & 27\% & 21\% & 54\% & \textbf{60\%} & 54\% \\
        Dump Bin Bigbin
            & 49\% & 68\% & 64\% & 83\% & 85\% & \textbf{86\%} \\
        Place Can Basket
            & 18\% & 1\% & 19\% & 41\% & 67\% & \textbf{68\%} \\
        Beat Block Hammer
            & 42\% & 56\% & 77\% & 43\% & 72\% & \textbf{78\%} \\
        Move Can Pot
            & 39\% & 22\% & 25\% & 58\% & 70\% & \textbf{86\%} \\
        \midrule
        Average
            & 35.1\% & 30.6\% & 35.8\% & 48.8\% & 59.5\% & \textbf{70.0\%} \\
        \bottomrule
    \end{tabularx}
\end{table*}

\subsection{General Manipulation on RoboTwin 2.0}

RoboTwin 2.0 contains bimanual tasks with randomized object placements, scene configurations, and manipulation geometries. In this benchmark, \method\ uses point-cloud observations and a trainable PointNet-style encoder. We treat RoboTwin 2.0 as a general manipulation benchmark rather than a memory-dependent benchmark. Most tasks are close to fully observable: once the current observation is known, the next action is mainly determined by local geometry, object pose, and contact configuration rather than distant historical phase. This benchmark tests whether full-history modeling remains competitive when memory is not the dominant bottleneck.

Figure~\ref{fig:robotwin_qualitative} further visualizes representative successful rollouts across the RoboTwin 2.0 tasks, showing that the policy maintains coherent action execution beyond memory-specific benchmarks.

Table~\ref{tab:robotwin} shows that \method\ is not only a memory-specialized policy.
It achieves the best average score and leads seven out of eight tasks.
The 95\% success rate on \emph{Place Cans Plasticbox} suggests that point-cloud state tokens plus acceleration refinement are especially effective when the task requires stable reach--grasp--place geometry under pose variation. On \emph{Hanging Mug}, the gain comes from maintaining handle pose and contact phase during the final hanging motion. \emph{Beat Block Hammer} benefits from smooth tool approach and impact control, where abrupt action changes can easily miss the block or destabilize the strike.
\emph{Move Can Pot} further reflects the advantage of coherent transport and release under object-pose variation. In \emph{Stack Bowls Three}, the gain is smaller because the task is already largely solved by strong present-state models; nevertheless, full-history encoding helps preserve the stage of the stacking sequence. \emph{Dump Bin Bigbin} and \emph{Place Can Basket} require robust transport and release, where second-order integration reduces abrupt local action changes.

The single task where \method\ underperforms DP3 is \emph{Put Bottles Dustbin}. This limitation is informative. The task is general fully observable and dominated by local geometry, object pose, and grasp-placement precision. A compact Markovian diffusion policy with a strong local geometric bias may generalize slightly better to position variation in such settings. This does not weaken the central claim; rather, it clarifies the boundary. When the present state is sufficient and the dominant challenge is local pose generalization, memory is less decisive. When observation aliasing is present, the advantage of full-history modeling becomes much larger.

\begin{figure*}[t]
    \centering
    \includegraphics[width=1\textwidth]{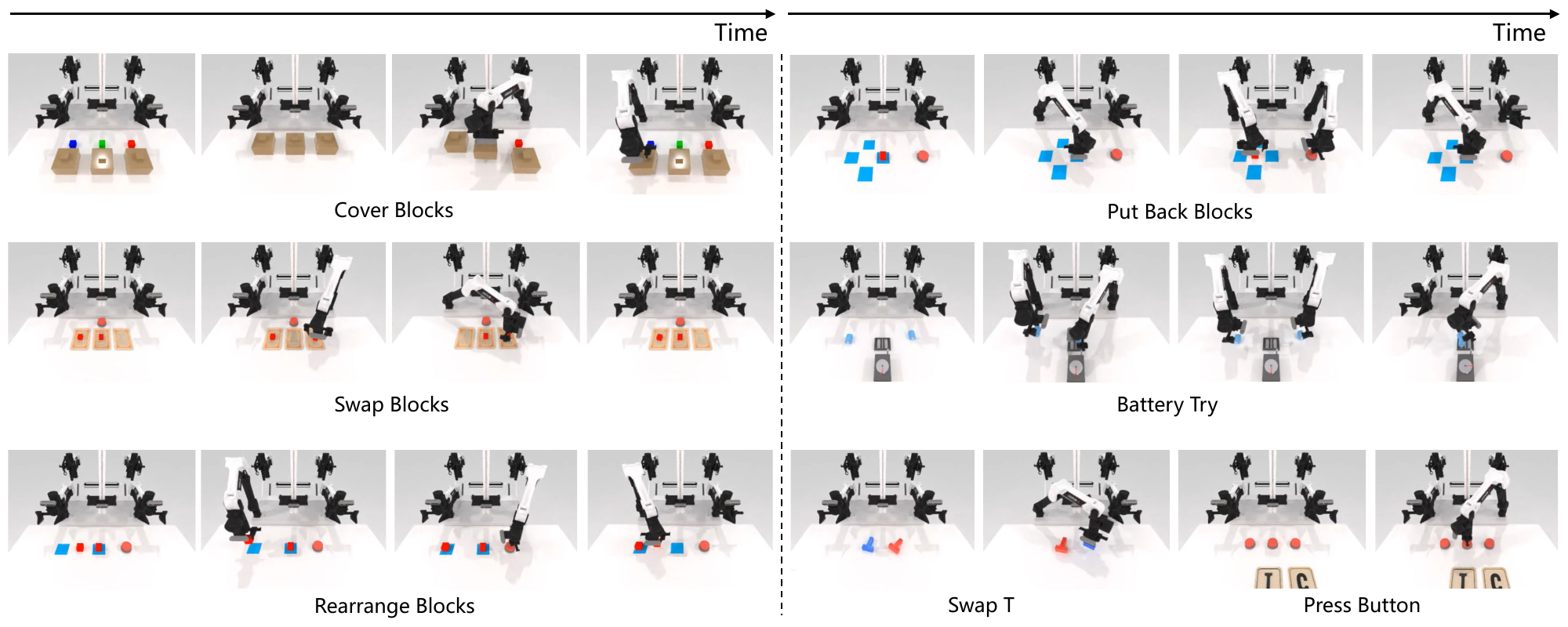}
    \vspace{-1mm}
    \caption{
    Qualitative rollouts on the seven RMBench memory-dependent manipulation tasks.
    Each row shows a representative temporal sequence.
    The tasks require the policy to retain hidden references, object order, attempted actions, or task phase over long horizons, making them difficult for Markovian observation policies.
    }
    \label{fig:rmbench_qualitative}
    \vspace{-2mm}
\end{figure*}

\begin{table*}[t]
    \centering
    \caption{RMBench memory-dependent success rates with 50 demonstrations and 100 evaluation trials. Baselines are officially reported results unless otherwise specified.}
    \label{tab:rmbench}
    \small
    \setlength{\tabcolsep}{1.9pt}
    \renewcommand{\arraystretch}{1.05}
    \begin{tabularx}{\textwidth}{@{}lYYYYYY@{}}
        \toprule
        Task
            & \makecell{DP\\(96.8M)}
            & \makecell{ACT\\(80.0M)}
            & \makecell{\(\pi_{0.5}\)\\(\(>3.3\)B)}
            & \makecell{X-VLA\\(0.9B)}
            & \makecell{Mem-0\\(\(>10\)B)}
            & \makecell{\method\\(0.3B)} \\
        \midrule
        Rearrange Blocks
            & 0\% & 29\% & 13\% & 13\% & 89\% & \textbf{98\%} \\
        Put Back Block
            & 0\% & 0\% & 11\% & 18\% & 90\% & \textbf{98\%} \\
        Swap Blocks
            & 11\% & 2\% & 24\% & 16\% & 67\% & \textbf{99\%} \\
        Swap T
            & 20\% & 2\% & 15\% & 3\% & 14\% & \textbf{93\%} \\
        Battery Try
            & 10\% & 19\% & 16\% & 26\% & 28\% & \textbf{29\%} \\
        Cover Blocks
            & 0\% & 0\% & 0\% & 2\% & 68\% & \textbf{96\%} \\
        Press Button
            & 0\% & 0\% & 0\% & 0\% & 0\% & \textbf{2\%} \\
        \midrule
        Average
            & 5.8\% & 7.4\% & 11.2\% & 11.1\% & 50.8\% & \textbf{73.6\%} \\
        \bottomrule
    \end{tabularx}
\end{table*}

\subsection{Memory-Dependent Manipulation on RMBench}

RMBench directly evaluates whether a robot remembers what happened before the current observation. The tasks include observing a hidden reference object, rearranging blocks after button-triggered phase changes, returning a block to its original pad, swapping block positions through an empty pad, swapping T-shaped objects with orientations, repeatedly trying battery insertions, covering and uncovering colored blocks, and pressing buttons according to digit tiles. These tasks differ in surface appearance, but mathematically they share one property: the correct policy is a function of path-dependent phase, not only present geometry.

Figure~\ref{fig:rmbench_qualitative} shows representative rollouts on the seven RMBench tasks.
The qualitative sequences illustrate the central challenge of the benchmark: the correct next action often depends on hidden task phase or earlier observations that are no longer visible in the current scene.
The average result is decisive. A 0.3B-parameter \method\ exceeds \(\pi_{0.5}\) by 62.4 points, X-VLA by 62.5 points, and Mem-0 by 22.8 points. Since Mem-0 is explicitly memory-aware, the comparison is more important than a simple Markovian baseline comparison. It indicates that full-history SSM encoding can be more effective than attaching explicit memory modules to a large VLA when precise temporal credit assignment is required.

\paragraph*{Rearrange Blocks}
This task requires the robot to move a middle block to a pad, press a button, and then move another block into the middle position. The challenge is not merely detecting blocks; it is knowing whether the button press has already occurred and which block is now active. Markovian policies confuse pre-button and post-button states. Mem-0 improves through sliding memory, but reported failures include redundant button presses. \method\ reaches 98\% because the SSM state evolves continuously through the full sequence, so button pressing is encoded as a phase transition rather than inferred from a fragile visual cue.

\paragraph*{Put Back Block}
The robot must remember the original pad after moving the block to the center and pressing the button. The current scene after the intermediate step does not by itself encode the source pad robustly. \method\ stores the initial source information in the trajectory-level hidden state, which explains the 98\% success compared with near-total collapse of Markovian baselines.

\begin{figure*}[t]
    \centering
    \includegraphics[width=0.98\textwidth]{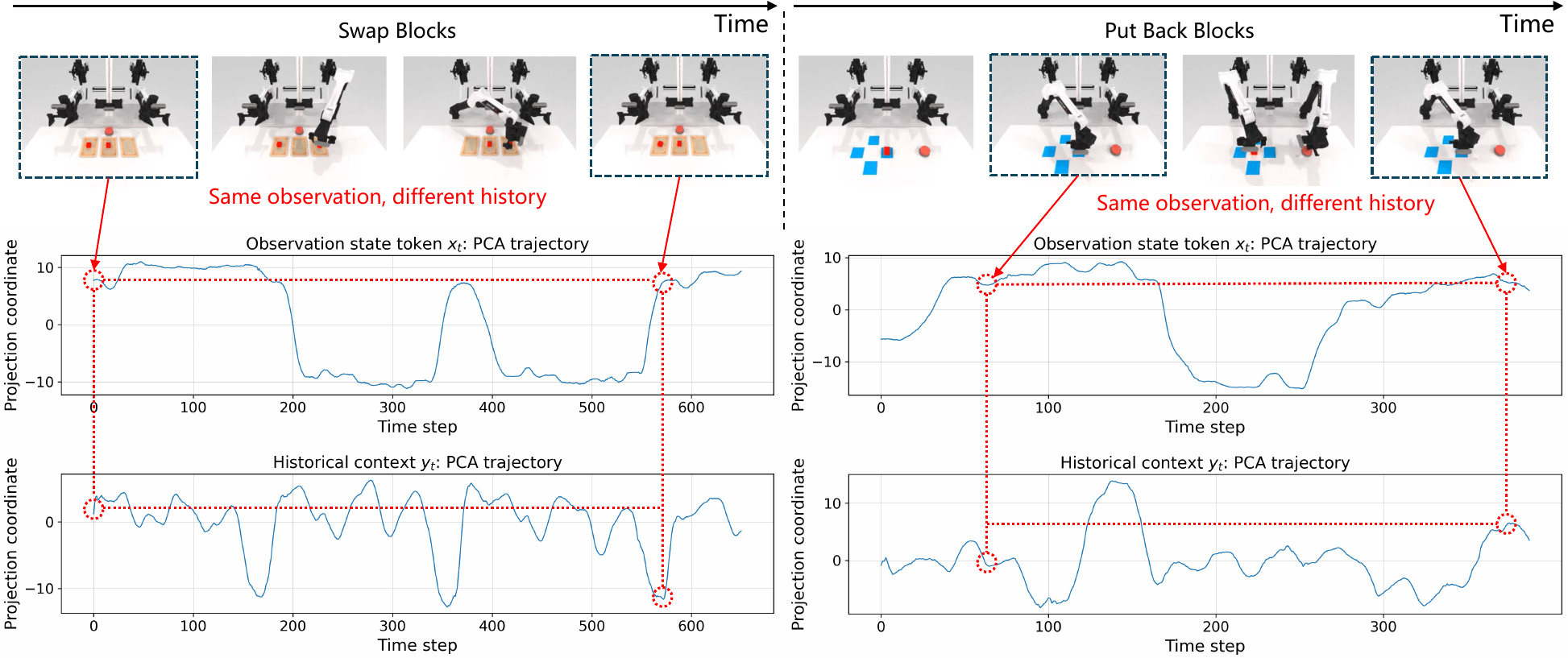}
    \vspace{-1mm}
    \caption{
    Representation visualization on memory-dependent RMBench tasks.
    Top rows show representative frames from Swap Blocks and Put Back Blocks, where visually similar observations are reached through different histories.
    Bottom plots show PCA trajectories of the instantaneous observation state token \(\bs{x}_t\) and the historical context \(\bs{y}_t\).
    Although the selected observations can be visually similar, the full-history context \(\bs{y}_t\) separates the corresponding task phases, indicating that the SSM history encoder carries information beyond the current observation.
    }
    \label{fig:history_visualization}
    \vspace{-2mm}
\end{figure*}

\paragraph*{Swap Blocks and Swap T}
Swapping requires maintaining a permutation state. For blocks, the empty pad acts as a temporary buffer; for T-shaped objects, both position and orientation must be tracked. This is a natural fit for recurrent state dynamics: the hidden state represents which objects have already been moved and what permutation remains. The 99\% and 93\% success rates suggest that one-token-per-time-step SSM memory captures this latent permutation more reliably than short-window visual policies.

\paragraph*{Cover Blocks}
The robot first covers blocks from left to right, then uncovers them in red--green--blue order, and finally returns covers. This task is strongly non-Markovian because the same cover and block geometry can correspond to multiple subtasks. Mem-0 failures are often classifier transition failures: the system cannot reliably detect whether an uncover subtask has ended. \method\ reaches 96\% because the subtask transition is represented inside the continuous historical state rather than delegated only to a downstream classifier.

\paragraph*{Battery Try}
All methods are low on Battery Try, and \method\ only slightly exceeds the best baseline. This is not primarily a memory failure. The task is physically brittle: two batteries with random orientations must be grasped and inserted into a dual-slot holder, and a slightly off-axis grasp or insertion can make a cylindrical battery slip away. Most failures in our rollouts are grasp or insertion failures rather than forgetting which insertion order has been tried. The small gap is therefore consistent with the benchmark physics: memory helps maintain attempted orders, but contact precision and simulator instability dominate.

\paragraph*{Press Button}
This task requires pressing the left and middle buttons according to two digit tiles and then pressing the right confirmation button. \method\ achieves a nonzero 2\% success while all reported baselines are zero. The task remains difficult because the present system lacks a dedicated OCR module and the dataset is too small for the policy to reliably learn the mapping from visual digits to press counts. The distinction from Markovian failure is important: even if a VLM recognizes the digits, a memoryless controller still cannot know how many presses have already been executed. \method\ provides the temporal counting substrate, but visual digit recognition and data coverage remain bottlenecks.

\subsection{Representation Visualization}
To further inspect whether the full-history encoder captures task-phase information, we visualize the temporal evolution of the instantaneous fused observation state token \(\bs{x}_t\) and the historical context \(\bs{y}_t\) on representative RMBench tasks, as shown in Figure~\ref{fig:history_visualization}. For each rollout, we project the high-dimensional features onto their first principal component and plot the resulting one-dimensional trajectory over control time. The highlighted pairs correspond to visually similar observations reached through different histories.

\paragraph*{Why Markovian and explicit-memory baselines fail}
The RMBench results show two different failure mechanisms. Markovian VLA baselines such as \(\pi_{0.5}\) and X-VLA are semantically strong, but they remain dynamically underdetermined when the current observation is aliased across different task phases. If the relevant event is no longer visible, increasing visual-language model scale does not by itself recover the missing state variable. Mem-0 is a stronger comparison because it explicitly introduces memory through anchor and sliding modules. Its improvement over Markovian VLA baselines confirms that memory is necessary. However, explicit memory modules can still fail when historical information must continuously modulate low-level action prediction rather than only be retrieved as a discrete cue. \method\ integrates memory into the recurrent policy state itself, so the hidden phase can evolve continuously and condition every action chunk.

\subsection{Real-World Experiments}

\begin{figure}[!t]
    \centering
    \includegraphics[width=0.98\columnwidth]{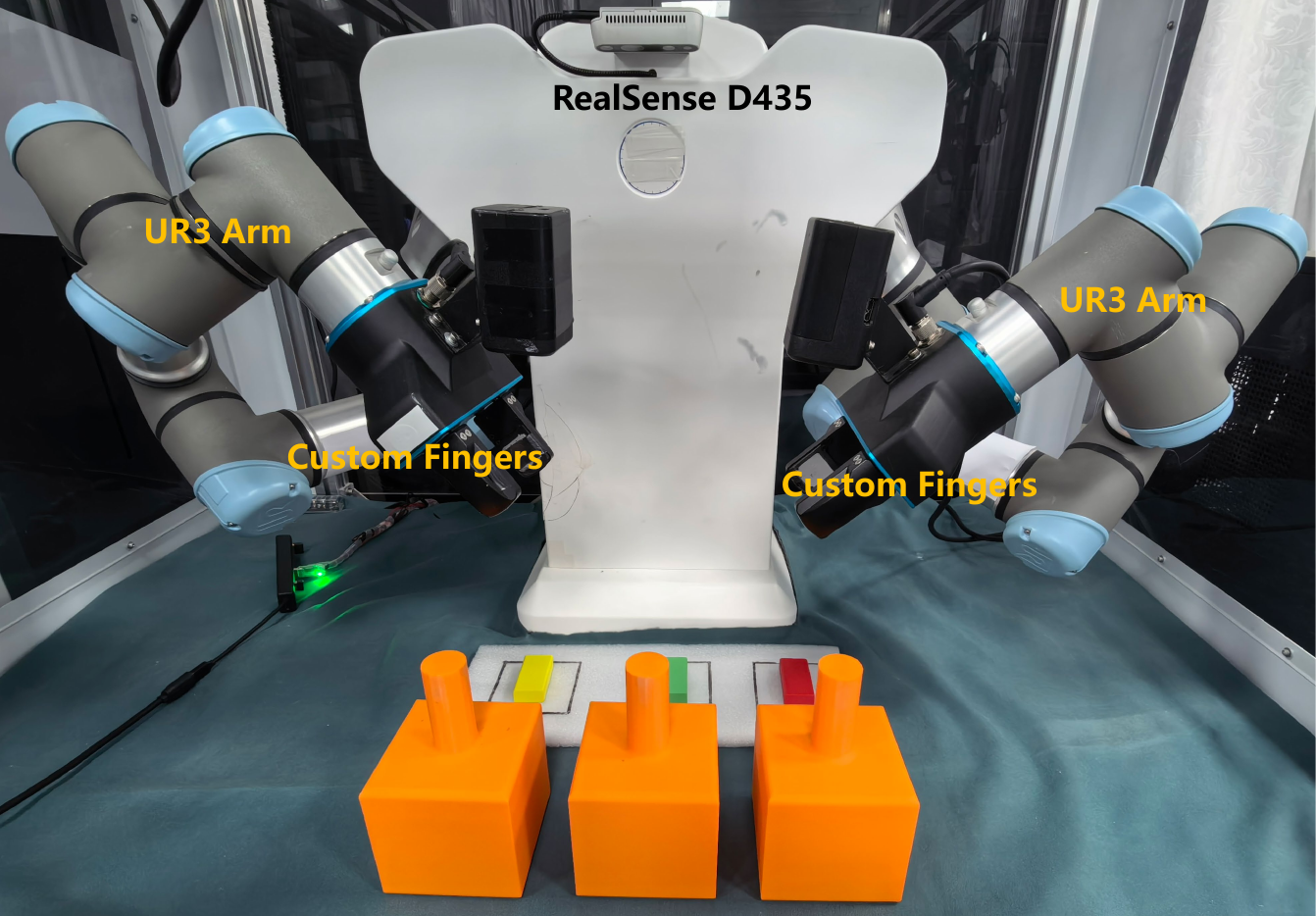}
    \vspace{-1mm}
    \caption{
    Real-world experimental setup. The platform consists of two UR3 arms, a single Intel RealSense D435 camera, and custom-made gripper fingers for dual-arm block manipulation.
    }
    \label{fig:real_setup}
    \vspace{-2mm}
\end{figure}

To evaluate transfer beyond simulation, we conduct real-world experiments on a dual-arm manipulation platform. As shown in Fig.~\ref{fig:real_setup}, the system consists of two UR3 arms and a single Intel RealSense D435 camera for visual observation. Both arms are equipped with custom-made gripper fingers designed for the block manipulation tasks considered in our experiments.

We further evaluate \method\ on four real-world manipulation tasks, each tested over 50 trials. The tasks are designed to separate geometric execution, historical phase memory, and action smoothness. We compare against \(\pi_{0.5}\), a strong Markovian VLA baseline. Both \method\ and \(\pi_{0.5}\) are trained and evaluated from a single camera view on the same robot platform. For \method, the visual token extractor is a frozen ResNet18 image backbone followed by trainable adapters and proprioceptive fusion; no point-cloud input is used in the real-world experiments because the D435 point cloud is noisy around the small blocks, covers, and contact regions. Videos are recorded for qualitative analysis.

\begin{figure*}[t]
    \centering
    \includegraphics[width=0.98\textwidth]{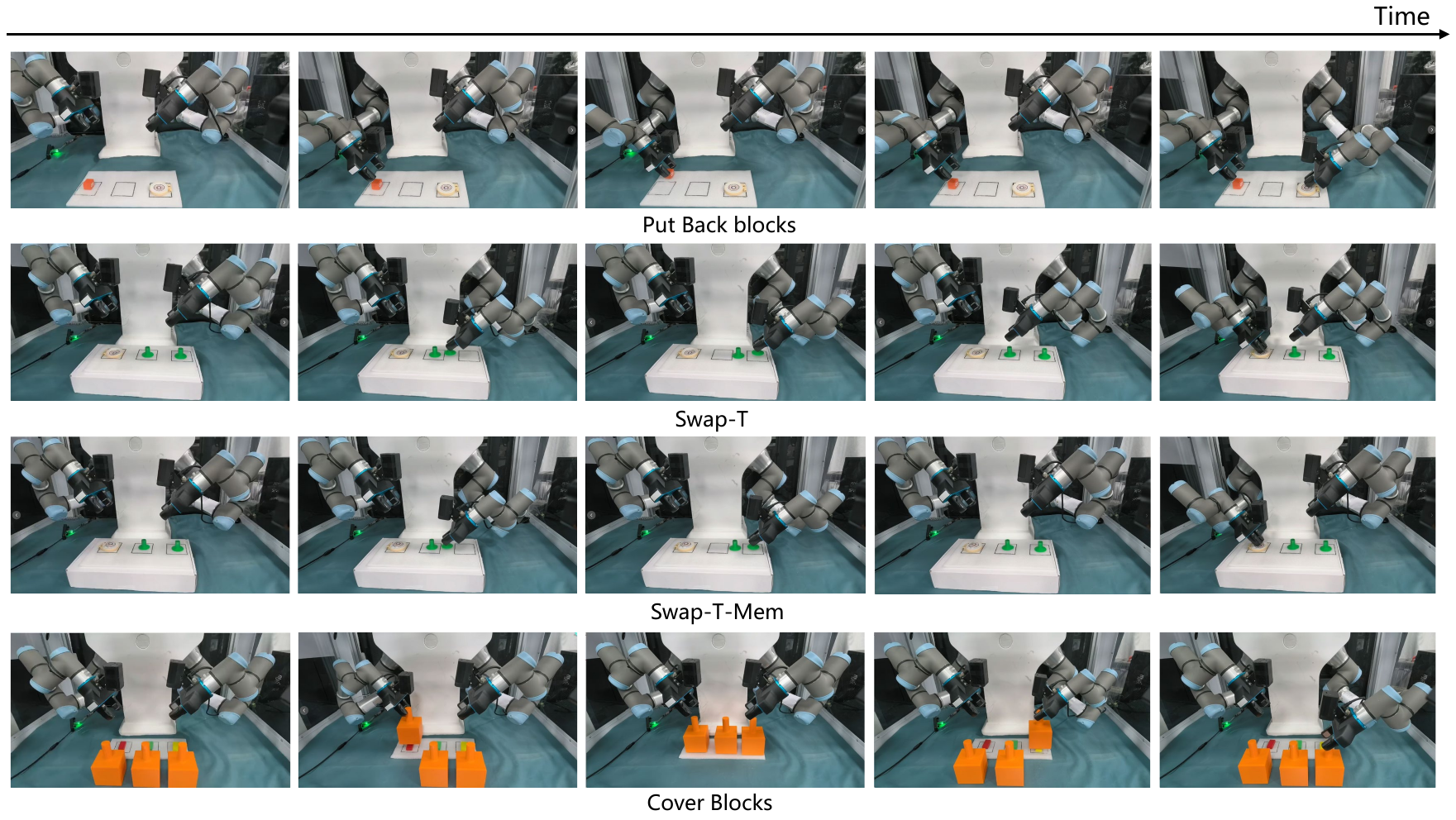}
    \vspace{-1mm}
    \caption{
    Real-world qualitative rollouts.
    We evaluate Chronos on four dual-arm tasks: Cover Blocks, Put Back Blocks, Swap-T-Mem, and Swap-T.
    The first three require historical phase or hidden-state memory, while Swap-T is designed as a non-memory control task.
    Chronos executes long-horizon bimanual manipulation smoothly from a single RGB camera view on a dual-UR3 platform with custom-made gripper fingers.
    }
    \label{fig:realworld_qualitative}
    \vspace{-2mm}
\end{figure*}
Figure~\ref{fig:realworld_qualitative} shows representative real-world rollouts.
The comparison between Swap-T and Swap-T-Mem is particularly important: the two tasks share similar object manipulation structure, but Swap-T-Mem deliberately introduces observation aliasing after the robot returns home, making historical memory necessary.
\paragraph*{Task definitions}
\emph{Put Back Blocks} requires the left arm to grasp a block, move it aside, grasp it again, return it to its original pose, and then let the right arm press a terminal button. This task requires remembering the original block state and the phase of the rearrangement. \emph{Swap T} is a non-memory control task: the right arm swaps two visually identical 3D-printed T-shaped objects through an intermediate buffer position, and the left arm then presses the terminal button. This task is deliberately designed to avoid observation aliasing, because after the second object is placed, the robot arm configuration still indicates task progress. \emph{Swap T-Mem} is the memory-dependent counterpart: after placing the second object, the right arm returns to its home pose, making the current observation visually similar to the initial frame. Correct execution therefore requires remembering that the first part of the swap has already happened. \emph{Cover Blocks} requires the robot to briefly observe three colored blocks arranged as red, green, and yellow, cover all of them, and then uncover the blocks in the order R--G--Y. Since the colors become hidden after covering, success requires persistent memory of the color-location association.

\begin{table}[t]
    \centering
    \caption{Real-world success rates over 50 trials per task. Both \method\ and \(\pi_{0.5}\) use a single camera view; Swap T is the non-memory control task.}
    \label{tab:real_world}
    \small
    \setlength{\tabcolsep}{4.2pt}
    \renewcommand{\arraystretch}{1.05}
    \begin{tabular}{@{}lcccc@{}}
        \toprule
        Task
            & Memory
            & \makecell{\(\pi_{0.5}\)\\(\(>3.3\)B)}
            & \makecell{\method\\(0.3B)}
            & Gain \\
        \midrule
        Put Back Blocks
            & Yes
            & 0/50 (0\%)
            & \textbf{49/50 (98\%)}
            & +98 pp \\
        Swap T
            & No
            & 14/50 (28\%)
            & \textbf{48/50 (96\%)}
            & +68 pp \\
        Swap T-Mem
            & Yes
            & 0/50 (0\%)
            & \textbf{49/50 (98\%)}
            & +98 pp \\
        Cover Blocks
            & Yes
            & 0/50 (0\%)
            & \textbf{10/50 (20\%)}
            & +20 pp \\
        \midrule
        All tasks
            & Mixed
            & 14/200 (7\%)
            & \textbf{156/200 (78\%)}
            & +71 pp \\
        Memory tasks
            & Yes
            & 0/150 (0\%)
            & \textbf{108/150 (72\%)}
            & +72 pp \\
        \bottomrule
    \end{tabular}
\end{table}

Table~\ref{tab:real_world} shows a clear separation between geometric competence and memory-dependent control. On the non-memory control task \emph{Swap T}, \(\pi_{0.5}\) obtains nonzero success, confirming that the baseline can sometimes solve real-world manipulation when the current observation still reveals task progress. However, its performance collapses to zero on the three memory-dependent tasks. In contrast, \method\ achieves 98\% on \emph{Put Back Blocks}, 98\% on \emph{Swap T-Mem}, and 20\% on the much harder \emph{Cover Blocks}. Averaged over the three memory-dependent tasks, \method\ reaches 72\% success while \(\pi_{0.5}\) remains at 0\%.

\paragraph*{Observation aliasing in Swap T-Mem}
The contrast between \emph{Swap T} and \emph{Swap T-Mem} is especially diagnostic. The two tasks share the same objects, geometry, and swap sequence, but \emph{Swap T-Mem} forces the robot arm to return to the home pose after placing the second object. As a result, a later observation becomes visually aliased with the initial observation. A Markovian VLA has no reliable state variable indicating whether the first half of the swap has already been completed. \method\ resolves this ambiguity because its hidden state evolves through the entire interaction history.

\paragraph*{Action smoothness}
The recorded videos show a qualitative difference in motion. The \(\pi_{0.5}\) rollouts often exhibit stop-and-go behavior, with visible frame-by-frame jitter resembling stop-motion execution. In contrast, \method\ produces smooth continuous trajectories in successful rollouts. We attribute this difference to the second-order action generator: the acceleration field is integrated through velocity before changing action position, which suppresses abrupt changes in action chunks. This real-world behavior is consistent with the motivation for using symplectic second-order dynamics rather than directly predicting actions or first-order velocities.

\paragraph*{Remaining difficulty in Cover Blocks}

Although \method\ is the only method that succeeds on \emph{Cover Blocks}, its real-world success rate is lower than its simulated RMBench counterpart.
This difference should not be interpreted as a general simulation-to-real gap of the proposed memory mechanism: the other three real-world tasks achieve success rates close to their simulated trends under the same policy framework.
A more direct explanation is the change of perception interface.
The simulated RMBench setting uses point-cloud input with a trainable PointNet-style encoder, which provides explicit geometric coordinates for color--location binding.
The real-world system instead uses a single RGB camera with a frozen ResNet18 backbone, because the D435 point cloud is noisy around small covered objects.
While effective for compact visual recognition, the convolutional and pooling structure of ResNet18 introduces translation-tolerant features, making it less sensitive to precise color--coordinate association after occlusion.
Most failures therefore arise from incorrect color-location binding after the blocks are covered, or from small grasping errors on the covers.
This is a perception-interface bottleneck rather than a contradiction of the full-history memory mechanism, and can be addressed by stronger 2D visual backbones such as DINOv2 or by using a higher-quality depth/LiDAR sensor, such as an Intel RealSense L515, to recover reliable point clouds and reuse the same point-cloud backbone as in RMBench.

\subsection{Failure Modes and Discussion}

The experiments support a simple theoretical message: history is not context; for many manipulation tasks, history is state.
When observation aliasing is present, a Markovian VLA can be semantically powerful yet dynamically underdetermined.
Larger backbones can recognize more objects, but they cannot resolve phase ambiguity if the relevant event is absent from the current observation.
A full-history SSM supplies the missing state variable by making the available past the native input to the policy dynamics.

The residual failures of \method\ differ from those of Markovian baselines.
Markovian failures are often phase failures: pressing a button at the wrong time, repeating an already completed action, or returning to the wrong target after an aliased observation.
In contrast, most \method\ failures are geometric or contact errors.
In Battery Try, the dominant failure is imprecise grasping or insertion that moves the battery into an unrecoverable pose.
In other RMBench tasks, failures are often slight placement offsets rather than loss of historical phase. This distinction suggests that the full-history mechanism addresses the intended memory bottleneck.

The real-world \emph{Swap T} and \emph{Swap T-Mem} comparison further isolates the role of history.
The two tasks share the same objects, geometry, and swap sequence, but \emph{Swap T-Mem} forces the robot arm to return to the home pose after placing the second object.
A later observation can then become visually aliased with the initial observation.
A Markovian VLA has no reliable state variable indicating whether the first half of the swap has already been completed.
\method\ resolves this ambiguity because its hidden state evolves through the entire interaction history.

The real-world \emph{Cover Blocks} result highlights a perception-interface issue rather than a failure of the memory mechanism.
The robot must first bind color identity to spatial location, preserve that hidden association after all blocks are covered, and finally uncover them in the correct R--G--Y order.
The simulated RMBench setting uses point-cloud input and a trainable PointNet-style encoder, whereas the real-world system uses a single RGB camera and a frozen ResNet18 backbone because the D435 point cloud is unreliable around small covered objects.
This 3D-to-2D change weakens explicit color--coordinate binding: ResNet18 provides translation-tolerant visual features, but the task requires precise memory of which color was at which location before occlusion.
The gap can therefore be reduced by stronger 2D visual backbones such as DINOv2, or by using a higher-quality depth/LiDAR camera such as an Intel RealSense L515 to obtain reliable point clouds and deploy the same point-cloud backbone as in RMBench.

The second message is that robot action generation benefits from higher-order structure.
Diffusion and flow matching are general and effective, but they operate through score or velocity fields.
In contrast, \method\ predicts acceleration and uses semi-implicit integration.
This does not mean that the proposed bridge is a full solution of a generic high-dimensional Schr\"odinger bridge problem.
It is a control-oriented Schr\"odinger-inspired projection: a coarse multimodal prior defines the bridge start, a cubic reference supplies a stable guide during training, and the learned acceleration field refines the action chunk.
This formulation gives a practical second-order inductive bias without claiming to solve the full density-transport problem.

Several practical extensions follow directly from the observed failure modes.
Digit-conditioned button pressing can benefit from a dedicated OCR or digit-recognition module.
Battery insertion and other contact-sensitive tasks can benefit from tactile or force feedback.
For fully observable tasks dominated by pose variation, stronger geometric augmentation and position generalization can further improve local policy robustness.
For real-world memory tasks with small occluded objects, improved perception is the most direct path: either a stronger 2D backbone for color--location binding, or a higher-quality depth/LiDAR camera that enables the same point-cloud pipeline used in simulation.

\section{Conclusion}

We presented \method, a physics-informed framework for non-Markovian long-horizon manipulation. The method combines full-history state space modeling, one state token per physical time step, a coarse multimodal action prior, and a second-order Schr\"odinger-inspired acceleration bridge derived from a Madelung--Kostin control projection. The results show that a 0.3B-parameter full-history policy can outperform much larger Markovian VLA models and a 10B-parameter memory VLA on memory-dependent manipulation. We believe this points toward a broader design principle for robot learning: the next generation of policies should scale not only parameters and data, but also the temporal and physical structure of action generation.

\section*{Acknowledgment}
This section is intentionally omitted for double-anonymous review.

\balance
\nocite{transformer,bert,vit,clip,pointnet,pointnetpp,pointtransformer,perceiver,bc,dagger,gail,ibc,ot,benamou2000,villani2009,pontryagin,neuralode,hamiltonian_nn,symplectic_ode,latentdiffusion,edm,robomimic,language_table,bc_z}
\bibliographystyle{IEEEtran}
\bibliography{chronos_refs}

@inproceedings{transformer,
  author    = {Ashish Vaswani and Noam Shazeer and Niki Parmar and Jakob Uszkoreit and Llion Jones and Aidan N. Gomez and Lukasz Kaiser and Illia Polosukhin},
  title     = {Attention Is All You Need},
  booktitle = {Advances in Neural Information Processing Systems},
  year      = {2017},
  url       = {https://arxiv.org/abs/1706.03762}
}

@inproceedings{bert,
  author    = {Jacob Devlin and Ming-Wei Chang and Kenton Lee and Kristina Toutanova},
  title     = {{BERT}: Pre-training of Deep Bidirectional Transformers for Language Understanding},
  booktitle = {Proceedings of the 2019 Conference of the North American Chapter of the Association for Computational Linguistics: Human Language Technologies},
  year      = {2019},
  url       = {https://arxiv.org/abs/1810.04805}
}

@inproceedings{vit,
  author    = {Alexey Dosovitskiy and Lucas Beyer and Alexander Kolesnikov and Dirk Weissenborn and Xiaohua Zhai and Thomas Unterthiner and Mostafa Dehghani and Matthias Minderer and Georg Heigold and Sylvain Gelly and Jakob Uszkoreit and Neil Houlsby},
  title     = {An Image is Worth 16x16 Words: Transformers for Image Recognition at Scale},
  booktitle = {International Conference on Learning Representations},
  year      = {2021},
  url       = {https://arxiv.org/abs/2010.11929}
}

@inproceedings{clip,
  author    = {Alec Radford and Jong Wook Kim and Chris Hallacy and Aditya Ramesh and Gabriel Goh and Sandhini Agarwal and Girish Sastry and Amanda Askell and Pamela Mishkin and Jack Clark and Gretchen Krueger and Ilya Sutskever},
  title     = {Learning Transferable Visual Models From Natural Language Supervision},
  booktitle = {Proceedings of the 38th International Conference on Machine Learning},
  series    = {Proceedings of Machine Learning Research},
  volume    = {139},
  year      = {2021},
  url       = {https://arxiv.org/abs/2103.00020}
}

@misc{openx,
  author        = {{Open X-Embodiment Collaboration}},
  title         = {Open X-Embodiment: Robotic Learning Datasets and {RT-X} Models},
  year          = {2023},
  eprint        = {2310.08864},
  archivePrefix = {arXiv},
  primaryClass  = {cs.RO},
  doi           = {10.48550/arXiv.2310.08864},
  url           = {https://arxiv.org/abs/2310.08864}
}

@inproceedings{openvla,
  author    = {Moo Jin Kim and Karl Pertsch and Siddharth Karamcheti and Ted Xiao and Ashwin Balakrishna and Suraj Nair and Rafael Rafailov and Ethan P. Foster and Pannag R. Sanketi and Quan Vuong and Thomas Kollar and Benjamin Burchfiel and Russ Tedrake and Dorsa Sadigh and Sergey Levine and Percy Liang and Chelsea Finn},
  title     = {{OpenVLA}: An Open-Source Vision-Language-Action Model},
  booktitle = {Proceedings of the 8th Conference on Robot Learning},
  series    = {Proceedings of Machine Learning Research},
  volume    = {270},
  pages     = {2679--2713},
  publisher = {PMLR},
  year      = {2025},
  url       = {https://proceedings.mlr.press/v270/kim25c.html},
  eprint    = {2406.09246},
  archivePrefix = {arXiv}
}

@inproceedings{pi0,
  author        = {Kevin Black and Noah Brown and Danny Driess and Adnan Esmail and Michael Equi and Chelsea Finn and Niccolo Fusai and Lachy Groom and Karol Hausman and Brian Ichter and Szymon Jakubczak and Tim Jones and Liyiming Ke and Sergey Levine and Adrian Li-Bell and Mohith Mothukuri and Suraj Nair and Karl Pertsch and Lucy Xiaoyang Shi and James Tanner and Quan Vuong and Anna Walling and Haohuan Wang and Ury Zhilinsky},
  title         = {{$\pi_0$}: A Vision-Language-Action Flow Model for General Robot Control},
  booktitle     = {Robotics: Science and Systems},
  year          = {2025},
  eprint        = {2410.24164},
  archivePrefix = {arXiv},
  primaryClass  = {cs.LG},
  doi           = {10.48550/arXiv.2410.24164},
  url           = {https://arxiv.org/abs/2410.24164}
}

@misc{pi05,
  author        = {{Physical Intelligence} and Kevin Black and Noah Brown and James Darpinian and Karan Dhabalia and Danny Driess and Adnan Esmail and Michael Equi and Chelsea Finn and Niccolo Fusai and Manuel Y. Galliker and Dibya Ghosh and Lachy Groom and Karol Hausman and Brian Ichter and Szymon Jakubczak and Tim Jones and Liyiming Ke and Devin LeBlanc and Sergey Levine and Adrian Li-Bell and Mohith Mothukuri and Suraj Nair and Karl Pertsch and Allen Z. Ren and Lucy Xiaoyang Shi and Laura Smith and Jost Tobias Springenberg and Kyle Stachowicz and James Tanner and Quan Vuong and Homer Walke and Anna Walling and Haohuan Wang and Lili Yu and Ury Zhilinsky},
  title         = {{$\pi_{0.5}$}: a Vision-Language-Action Model with Open-World Generalization},
  year          = {2025},
  eprint        = {2504.16054},
  archivePrefix = {arXiv},
  primaryClass  = {cs.LG},
  doi           = {10.48550/arXiv.2504.16054},
  url           = {https://arxiv.org/abs/2504.16054}
}

@misc{pi06,
  author        = {{Physical Intelligence} and Ali Amin and Raichelle Aniceto and Ashwin Balakrishna and Kevin Black and Ken Conley and Grace Connors and James Darpinian and Karan Dhabalia and Jared DiCarlo and Danny Driess and Michael Equi and Adnan Esmail and Yunhao Fang and Chelsea Finn and Catherine Glossop and Thomas Godden and Ivan Goryachev and Lachy Groom and Hunter Hancock and Karol Hausman and Gashon Hussein and Brian Ichter and Szymon Jakubczak and Rowan Jen and Tim Jones and Ben Katz and Liyiming Ke and Chandra Kuchi and Marinda Lamb and Devin LeBlanc and Sergey Levine and Adrian Li-Bell and Yao Lu and Vishnu Mano and Mohith Mothukuri and Suraj Nair and Karl Pertsch and Allen Z. Ren and Charvi Sharma and Lucy Xiaoyang Shi and Laura Smith and Jost Tobias Springenberg and Kyle Stachowicz and Will Stoeckle and Alex Swerdlow and James Tanner and Marcel Torne and Quan Vuong and Anna Walling and Haohuan Wang and Blake Williams and Sukwon Yoo and Lili Yu and Ury Zhilinsky and Zhiyuan Zhou},
  title         = {{$\pi^{*}_{0.6}$}: a {VLA} That Learns From Experience},
  year          = {2025},
  eprint        = {2511.14759},
  archivePrefix = {arXiv},
  primaryClass  = {cs.LG},
  doi           = {10.48550/arXiv.2511.14759},
  url           = {https://arxiv.org/abs/2511.14759}
}

@misc{pi07,
  author        = {{Physical Intelligence} and Bo Ai and Ali Amin and Raichelle Aniceto and Ashwin Balakrishna and Greg Balke and Kevin Black and George Bokinsky and Shihao Cao and Thomas Charbonnier and Vedant Choudhary and Foster Collins and Ken Conley and Grace Connors and James Darpinian and Karan Dhabalia and Maitrayee Dhaka and Jared DiCarlo and Danny Driess and Michael Equi and Adnan Esmail and Yunhao Fang and Chelsea Finn and Catherine Glossop and Thomas Godden and Ivan Goryachev and Lachlan Groom and Haroun Habeeb and Hunter Hancock and Karol Hausman and Gashon Hussein and Victor Hwang and Brian Ichter and Connor Jacobsen and Szymon Jakubczak and Rowan Jen and Tim Jones and Gregg Kammerer and Ben Katz and Liyiming Ke and Mairbek Khadikov and Chandra Kuchi and Marinda Lamb and Devin LeBlanc and Brendon LeCount and Sergey Levine and Xinyu Li and Adrian Li-Bell and Vladislav Lialin and Zhonglin Liang and Wallace Lim and Yao Lu and Enyu Luo and Vishnu Mano and Nandan Marwaha and Aikys Mongush and Liam Murphy and Suraj Nair and Tyler Patterson and Karl Pertsch and Allen Z. Ren and Gavin Schelske and Charvi Sharma and Baifeng Shi and Lucy Xiaoyang Shi and Laura Smith and Jost Tobias Springenberg and Kyle Stachowicz and Will Stoeckle and Jiaming Tang and Jimmy Tanner and Shalom Tekeste and Marcel Torne and Kyle Vedder and Quan Vuong and Anna Walling and Haohuan Wang and Jason Wang and Xudong Wang and Chris Whalen and Samuel Whitmore and Blake Williams and Charles Xu and Sukwon Yoo and Lili Yu and Wuming Zhang and Zhuoyang Zhang and Ury Zhilinsky},
  title         = {{$\pi_{0.7}$}: a Steerable Generalist Robotic Foundation Model with Emergent Capabilities},
  year          = {2026},
  eprint        = {2604.15483},
  archivePrefix = {arXiv},
  primaryClass  = {cs.LG},
  doi           = {10.48550/arXiv.2604.15483},
  url           = {https://arxiv.org/abs/2604.15483}
}

@misc{rdt1b,
  author        = {Songming Liu and Lingxuan Wu and Bangguo Li and Hengkai Tan and Huayu Chen and Zhengyi Wang and Ke Xu and Hang Su and Jun Zhu},
  title         = {{RDT-1B}: a Diffusion Foundation Model for Bimanual Manipulation},
  year          = {2024},
  eprint        = {2410.07864},
  archivePrefix = {arXiv},
  primaryClass  = {cs.RO},
  doi           = {10.48550/arXiv.2410.07864},
  url           = {https://arxiv.org/abs/2410.07864}
}

@misc{rdt2,
  author        = {Songming Liu and Bangguo Li and Kai Ma and Lingxuan Wu and Hengkai Tan and Xiao Ouyang and Hang Su and Jun Zhu},
  title         = {{RDT2}: Exploring the Scaling Limit of {UMI} Data Towards Zero-Shot Cross-Embodiment Generalization},
  year          = {2026},
  eprint        = {2602.03310},
  archivePrefix = {arXiv},
  primaryClass  = {cs.RO},
  doi           = {10.48550/arXiv.2602.03310},
  url           = {https://arxiv.org/abs/2602.03310}
}

@inproceedings{xvla,
  author        = {Jinliang Zheng and Jianxiong Li and Zhihao Wang and Dongxiu Liu and Xirui Kang and Yuchun Feng and Yinan Zheng and Jiayin Zou and Yilun Chen and Jia Zeng and Tai Wang and Ya-Qin Zhang and Jingjing Liu and Xianyuan Zhan},
  title         = {{X-VLA}: Soft-Prompted Transformer as Scalable Cross-Embodiment Vision-Language-Action Model},
  booktitle     = {International Conference on Learning Representations},
  year          = {2026},
  url           = {https://openreview.net/forum?id=kt51kZH4aG},
  eprint        = {2510.10274},
  archivePrefix = {arXiv},
  doi           = {10.48550/arXiv.2510.10274}
}

@misc{motus,
  author        = {Hongzhe Bi and Hengkai Tan and Shenghao Xie and Zeyuan Wang and Shuhe Huang and Haitian Liu and Ruowen Zhao and Yao Feng and Chendong Xiang and Yinze Rong and Hongyan Zhao and Hanyu Liu and Zhizhong Su and Lei Ma and Hang Su and Jun Zhu},
  title         = {Motus: A Unified Latent Action World Model},
  year          = {2025},
  eprint        = {2512.13030},
  archivePrefix = {arXiv},
  primaryClass  = {cs.CV},
  doi           = {10.48550/arXiv.2512.13030},
  url           = {https://arxiv.org/abs/2512.13030}
}

@inproceedings{act,
  author    = {Tony Z. Zhao and Vikash Kumar and Sergey Levine and Chelsea Finn},
  title     = {Learning Fine-Grained Bimanual Manipulation with Low-Cost Hardware},
  booktitle = {Robotics: Science and Systems},
  year      = {2023},
  url       = {https://arxiv.org/abs/2304.13705}
}

@misc{aloha,
  author       = {Tony Z. Zhao and Vikash Kumar and Sergey Levine and Chelsea Finn},
  title        = {{ALOHA}: A Low-Cost Open-Source Hardware System for Bimanual Teleoperation},
  year         = {2023},
  howpublished = {Project website},
  url          = {https://tonyzhaozh.github.io/aloha/},
  note         = {Official project page for the ALOHA hardware system used in Zhao et al., ``Learning Fine-Grained Bimanual Manipulation with Low-Cost Hardware.''}
}

@article{diffusionpolicy,
  author  = {Cheng Chi and Zhenjia Xu and Siyuan Feng and Eric Cousineau and Yilun Du and Benjamin Burchfiel and Russ Tedrake and Shuran Song},
  title   = {Diffusion Policy: Visuomotor Policy Learning via Action Diffusion},
  journal = {The International Journal of Robotics Research},
  year    = {2025},
  doi     = {10.1177/02783649241273668},
  url     = {https://arxiv.org/abs/2303.04137}
}

@misc{dp3,
  author        = {Yanjie Ze and Gu Zhang and Kangning Zhang and Chenyuan Hu and Muhan Wang and Huazhe Xu},
  title         = {3D Diffusion Policy: Generalizable Visuomotor Policy Learning via Simple 3D Representations},
  year          = {2024},
  eprint        = {2403.03954},
  archivePrefix = {arXiv},
  primaryClass  = {cs.RO},
  doi           = {10.48550/arXiv.2403.03954},
  url           = {https://arxiv.org/abs/2403.03954}
}

@inproceedings{consistencypolicy,
  author    = {Aaditya Prasad and Kevin Lin and Jimmy Wu and Linqi Zhou and Jeannette Bohg},
  title     = {Consistency Policy: Accelerated Visuomotor Policies via Consistency Distillation},
  booktitle = {Robotics: Science and Systems},
  year      = {2024},
  eprint    = {2405.07503},
  archivePrefix = {arXiv},
  url       = {https://arxiv.org/abs/2405.07503}
}

@inproceedings{flowmatching,
  author    = {Yaron Lipman and Ricky T. Q. Chen and Heli Ben-Hamu and Maximilian Nickel and Matt Le},
  title     = {Flow Matching for Generative Modeling},
  booktitle = {International Conference on Learning Representations},
  year      = {2023},
  url       = {https://arxiv.org/abs/2210.02747}
}

@inproceedings{rectifiedflow,
  author    = {Xingchao Liu and Chengyue Gong and Qiang Liu},
  title     = {Flow Straight and Fast: Learning to Generate and Transfer Data with Rectified Flow},
  booktitle = {International Conference on Learning Representations},
  year      = {2023},
  url       = {https://arxiv.org/abs/2209.03003}
}

@misc{flowpolicy,
  author        = {Qinglun Zhang and Zhen Liu and Haoqiang Fan and Guanghui Liu and Bing Zeng and Shuaicheng Liu},
  title         = {FlowPolicy: Enabling Fast and Robust 3D Flow-based Policy via Consistency Flow Matching for Robot Manipulation},
  year          = {2024},
  eprint        = {2412.04987},
  archivePrefix = {arXiv},
  primaryClass  = {cs.RO},
  doi           = {10.48550/arXiv.2412.04987},
  url           = {https://arxiv.org/abs/2412.04987}
}

@misc{actionflow,
  author        = {Niklas Funk and Julen Urain and Joao Carvalho and Vignesh Prasad and Georgia Chalvatzaki and Jan Peters},
  title         = {ActionFlow: Equivariant, Accurate, and Efficient Policies with Spatially Symmetric Flow Matching},
  year          = {2024},
  eprint        = {2409.04576},
  archivePrefix = {arXiv},
  primaryClass  = {cs.RO},
  doi           = {10.48550/arXiv.2409.04576},
  url           = {https://arxiv.org/abs/2409.04576}
}

@inproceedings{imle,
  author    = {Ke Li and Jitendra Malik},
  title     = {Implicit Maximum Likelihood Estimation},
  booktitle = {International Conference on Learning Representations},
  year      = {2018},
  url       = {https://arxiv.org/abs/1809.09087}
}

@inproceedings{imlepolicy,
  author    = {Krishan Rana and Robert Lee and David Pershouse and Niko Suenderhauf},
  title     = {{IMLE} Policy: Fast and Sample Efficient Visuomotor Policy Learning via Implicit Maximum Likelihood Estimation},
  booktitle = {Robotics: Science and Systems},
  year      = {2025},
  eprint    = {2502.12371},
  archivePrefix = {arXiv},
  doi       = {10.15607/RSS.2025.XXI.158},
  url       = {https://arxiv.org/abs/2502.12371}
}

@inproceedings{s4,
  author        = {Albert Gu and Karan Goel and Christopher R{\'e}},
  title         = {Efficiently Modeling Long Sequences with Structured State Spaces},
  booktitle     = {International Conference on Learning Representations},
  year          = {2022},
  eprint        = {2111.00396},
  archivePrefix = {arXiv},
  primaryClass  = {cs.LG},
  url           = {https://arxiv.org/abs/2111.00396}
}

@misc{mamba,
  author        = {Albert Gu and Tri Dao},
  title         = {Mamba: Linear-Time Sequence Modeling with Selective State Spaces},
  year          = {2023},
  eprint        = {2312.00752},
  archivePrefix = {arXiv},
  primaryClass  = {cs.LG},
  doi           = {10.48550/arXiv.2312.00752},
  url           = {https://arxiv.org/abs/2312.00752}
}

@inproceedings{mamba2,
  author        = {Tri Dao and Albert Gu},
  title         = {Transformers are {SSMs}: Generalized Models and Efficient Algorithms Through Structured State Space Duality},
  booktitle     = {International Conference on Machine Learning},
  year          = {2024},
  eprint        = {2405.21060},
  archivePrefix = {arXiv},
  primaryClass  = {cs.LG},
  url           = {https://arxiv.org/abs/2405.21060}
}

@article{mtil,
  author  = {Yulin Zhou and Yuankai Lin and Fanzhe Peng and Jiahui Chen and Kaiji Huang and Hua Yang and Zhouping Yin},
  title   = {{MTIL}: Encoding Full History with Mamba for Temporal Imitation Learning},
  journal = {IEEE Robotics and Automation Letters},
  year    = {2025},
  doi     = {10.1109/LRA.2025.3615520},
  url     = {https://arxiv.org/abs/2505.12410}
}

@misc{rmbench,
  author        = {Tianxing Chen and Yuran Wang and Mingleyang Li and Yan Qin and Hao Shi and Zixuan Li and Yifan Hu and Yingsheng Zhang and Kaixuan Wang and Yue Chen and Hongcheng Wang and Renjing Xu and Ruihai Wu and Yao Mu and Yaodong Yang and Hao Dong and Ping Luo},
  title         = {{RMBench}: Memory-Dependent Robotic Manipulation Benchmark with Insights into Policy Design},
  year          = {2026},
  eprint        = {2603.01229},
  archivePrefix = {arXiv},
  primaryClass  = {cs.RO},
  doi           = {10.48550/arXiv.2603.01229},
  url           = {https://arxiv.org/abs/2603.01229}
}

@misc{robomme,
  author        = {Yinpei Dai and Hongze Fu and Jayjun Lee and Yuejiang Liu and Haoran Zhang and Jianing Yang and Chelsea Finn and Nima Fazeli and Joyce Chai},
  title         = {{RoboMME}: Benchmarking and Understanding Memory for Robotic Generalist Policies},
  year          = {2026},
  eprint        = {2603.04639},
  archivePrefix = {arXiv},
  primaryClass  = {cs.RO},
  doi           = {10.48550/arXiv.2603.04639},
  url           = {https://arxiv.org/abs/2603.04639}
}

@misc{memoryvla,
  author        = {Hao Shi and Bin Xie and Yingfei Liu and Lin Sun and Fengrong Liu and Tiancai Wang and Erjin Zhou and Haoqiang Fan and Xiangyu Zhang and Gao Huang},
  title         = {{MemoryVLA}: Perceptual-Cognitive Memory in Vision-Language-Action Models for Robotic Manipulation},
  year          = {2025},
  eprint        = {2508.19236},
  archivePrefix = {arXiv},
  primaryClass  = {cs.RO},
  doi           = {10.48550/arXiv.2508.19236},
  url           = {https://arxiv.org/abs/2508.19236}
}

@misc{rememvla,
  author        = {Hang Li and Fengyi Shen and Dong Chen and Liudi Yang and Xudong Wang and Jinkui Shi and Zhenshan Bing and Ziyuan Liu and Alois Knoll},
  title         = {{ReMem-VLA}: Empowering Vision-Language-Action Model with Memory via Dual-Level Recurrent Queries},
  year          = {2026},
  eprint        = {2603.12942},
  archivePrefix = {arXiv},
  primaryClass  = {cs.RO},
  doi           = {10.48550/arXiv.2603.12942},
  url           = {https://arxiv.org/abs/2603.12942}
}

@misc{memoact,
  author        = {Liufan Tan and Jiale Li and Gangshan Jing},
  title         = {{MemoAct}: Atkinson--Shiffrin-Inspired Memory-Augmented Visuomotor Policy for Robotic Manipulation},
  year          = {2026},
  eprint        = {2603.18494},
  archivePrefix = {arXiv},
  primaryClass  = {cs.RO},
  doi           = {10.48550/arXiv.2603.18494},
  url           = {https://arxiv.org/abs/2603.18494}
}

@inproceedings{pointnet,
  author    = {Charles R. Qi and Hao Su and Kaichun Mo and Leonidas J. Guibas},
  title     = {PointNet: Deep Learning on Point Sets for 3D Classification and Segmentation},
  booktitle = {IEEE Conference on Computer Vision and Pattern Recognition},
  year      = {2017},
  url       = {https://arxiv.org/abs/1612.00593}
}

@inproceedings{pointnetpp,
  author    = {Charles R. Qi and Li Yi and Hao Su and Leonidas J. Guibas},
  title     = {PointNet++: Deep Hierarchical Feature Learning on Point Sets in a Metric Space},
  booktitle = {Advances in Neural Information Processing Systems},
  year      = {2017},
  url       = {https://arxiv.org/abs/1706.02413}
}

@inproceedings{pointtransformer,
  author    = {Hengshuang Zhao and Li Jiang and Jiaya Jia and Philip H. S. Torr and Vladlen Koltun},
  title     = {Point Transformer},
  booktitle = {IEEE International Conference on Computer Vision},
  year      = {2021},
  url       = {https://arxiv.org/abs/2012.09164}
}

@inproceedings{perceiver,
  author    = {Andrew Jaegle and Felix Gimeno and Andrew Brock and Andrew Zisserman and Oriol Vinyals and Joao Carreira},
  title     = {Perceiver: General Perception with Iterative Attention},
  booktitle = {Proceedings of the 38th International Conference on Machine Learning},
  series    = {Proceedings of Machine Learning Research},
  volume    = {139},
  year      = {2021},
  url       = {https://arxiv.org/abs/2103.03206}
}

@article{schrodinger1931,
  author  = {Erwin Schr{\"o}dinger},
  title   = {{\"U}ber die Umkehrung der Naturgesetze},
  journal = {Sitzungsberichte der Preussischen Akademie der Wissenschaften},
  pages   = {144--153},
  year    = {1931}
}

@article{madelung1927,
  author  = {Erwin Madelung},
  title   = {Quantentheorie in hydrodynamischer Form},
  journal = {Zeitschrift f{\"u}r Physik},
  volume  = {40},
  number  = {3--4},
  pages   = {322--326},
  year    = {1927},
  doi     = {10.1007/BF01400372}
}

@article{bohm1952,
  author  = {David Bohm},
  title   = {A Suggested Interpretation of the Quantum Theory in Terms of Hidden Variables. I},
  journal = {Physical Review},
  volume  = {85},
  number  = {2},
  pages   = {166--179},
  year    = {1952},
  doi     = {10.1103/PhysRev.85.166}
}

@book{holland1993,
  author    = {Peter R. Holland},
  title     = {The Quantum Theory of Motion: An Account of the de Broglie--Bohm Causal Interpretation of Quantum Mechanics},
  publisher = {Cambridge University Press},
  year      = {1993},
  doi       = {10.1017/CBO9780511622687}
}

@article{kostin1972,
  author  = {M. D. Kostin},
  title   = {On the Schr{\"o}dinger-Langevin Equation},
  journal = {The Journal of Chemical Physics},
  volume  = {57},
  number  = {9},
  pages   = {3589--3591},
  year    = {1972},
  doi     = {10.1063/1.1678812}
}

@article{leonard2014,
  author  = {Christian L{\'e}onard},
  title   = {A Survey of the Schr{\"o}dinger Problem and Some of Its Connections with Optimal Transport},
  journal = {Discrete and Continuous Dynamical Systems A},
  volume  = {34},
  number  = {4},
  pages   = {1533--1574},
  year    = {2014},
  doi     = {10.3934/dcds.2014.34.1533}
}

@article{chen2016sb,
  author  = {Yongxin Chen and Tryphon T. Georgiou and Michele Pavon},
  title   = {Optimal Transport Over a Linear Dynamical System},
  journal = {IEEE Transactions on Automatic Control},
  volume  = {62},
  number  = {5},
  pages   = {2137--2152},
  year    = {2017},
  doi     = {10.1109/TAC.2016.2604763}
}

@inproceedings{debortoli2021dsb,
  author    = {Valentin De Bortoli and James Thornton and Jeremy Heng and Arnaud Doucet},
  title     = {Diffusion Schr{\"o}dinger Bridge with Applications to Score-Based Generative Modeling},
  booktitle = {Advances in Neural Information Processing Systems},
  year      = {2021},
  url       = {https://arxiv.org/abs/2106.01357}
}

@inproceedings{vargas2021sb,
  author    = {Francisco Vargas and Pierre Thodoroff and Austen Lamacraft and Neil Lawrence},
  title     = {Solving Schr{\"o}dinger Bridges via Maximum Likelihood},
  booktitle = {International Conference on Learning Representations},
  year      = {2021},
  url       = {https://arxiv.org/abs/2106.02081}
}

@inproceedings{ddpm,
  author    = {Jonathan Ho and Ajay Jain and Pieter Abbeel},
  title     = {Denoising Diffusion Probabilistic Models},
  booktitle = {Advances in Neural Information Processing Systems},
  year      = {2020},
  url       = {https://arxiv.org/abs/2006.11239}
}

@inproceedings{scorebased,
  author    = {Yang Song and Jascha Sohl-Dickstein and Diederik P. Kingma and Abhishek Kumar and Stefano Ermon and Ben Poole},
  title     = {Score-Based Generative Modeling Through Stochastic Differential Equations},
  booktitle = {International Conference on Learning Representations},
  year      = {2021},
  url       = {https://arxiv.org/abs/2011.13456}
}

@inproceedings{latentdiffusion,
  author    = {Robin Rombach and Andreas Blattmann and Dominik Lorenz and Patrick Esser and Bj{\"o}rn Ommer},
  title     = {High-Resolution Image Synthesis with Latent Diffusion Models},
  booktitle = {IEEE/CVF Conference on Computer Vision and Pattern Recognition},
  year      = {2022},
  url       = {https://arxiv.org/abs/2112.10752}
}

@inproceedings{edm,
  author    = {Tero Karras and Miika Aittala and Timo Aila and Samuli Laine},
  title     = {Elucidating the Design Space of Diffusion-Based Generative Models},
  booktitle = {Advances in Neural Information Processing Systems},
  year      = {2022},
  url       = {https://arxiv.org/abs/2206.00364}
}

@book{ot,
  author    = {C{\'e}dric Villani},
  title     = {Topics in Optimal Transportation},
  publisher = {American Mathematical Society},
  year      = {2003},
  doi       = {10.1090/gsm/058}
}

@book{villani2009,
  author    = {C{\'e}dric Villani},
  title     = {Optimal Transport: Old and New},
  publisher = {Springer},
  year      = {2009},
  doi       = {10.1007/978-3-540-71050-9}
}

@article{benamou2000,
  author  = {Jean-David Benamou and Yann Brenier},
  title   = {A Computational Fluid Mechanics Solution to the Monge-Kantorovich Mass Transfer Problem},
  journal = {Numerische Mathematik},
  volume  = {84},
  pages   = {375--393},
  year    = {2000},
  doi     = {10.1007/s002110050002}
}

@book{pontryagin,
  author    = {Lev S. Pontryagin and Vladimir G. Boltyanskii and Revaz V. Gamkrelidze and Evgenii F. Mishchenko},
  title     = {The Mathematical Theory of Optimal Processes},
  publisher = {Interscience Publishers},
  year      = {1962}
}

@inproceedings{neuralode,
  author    = {Ricky T. Q. Chen and Yulia Rubanova and Jesse Bettencourt and David Duvenaud},
  title     = {Neural Ordinary Differential Equations},
  booktitle = {Advances in Neural Information Processing Systems},
  year      = {2018},
  url       = {https://arxiv.org/abs/1806.07366}
}

@inproceedings{hamiltonian_nn,
  author    = {Samuel Greydanus and Misko Dzamba and Jason Yosinski},
  title     = {Hamiltonian Neural Networks},
  booktitle = {Advances in Neural Information Processing Systems},
  year      = {2019},
  url       = {https://arxiv.org/abs/1906.01563}
}

@article{symplectic_ode,
  author  = {Pengzhan Jin and Zhen Zhang and Aiqing Zhu and Yifa Tang and George Em Karniadakis},
  title   = {{SympNets}: Intrinsic Structure-Preserving Symplectic Networks for Identifying Hamiltonian Systems},
  journal = {Neural Networks},
  volume  = {132},
  pages   = {166--179},
  year    = {2020},
  doi     = {10.1016/j.neunet.2020.08.017},
  url     = {https://arxiv.org/abs/2001.03750}
}

@article{bc,
  author  = {Dean A. Pomerleau},
  title   = {{ALVINN}: An Autonomous Land Vehicle in a Neural Network},
  journal = {Advances in Neural Information Processing Systems},
  year    = {1989}
}

@inproceedings{dagger,
  author    = {St{\'e}phane Ross and Geoffrey J. Gordon and J. Andrew Bagnell},
  title     = {A Reduction of Imitation Learning and Structured Prediction to No-Regret Online Learning},
  booktitle = {International Conference on Artificial Intelligence and Statistics},
  year      = {2011}
}

@inproceedings{gail,
  author    = {Jonathan Ho and Stefano Ermon},
  title     = {Generative Adversarial Imitation Learning},
  booktitle = {Advances in Neural Information Processing Systems},
  year      = {2016},
  url       = {https://arxiv.org/abs/1606.03476}
}

@inproceedings{ibc,
  author    = {Pete Florence and Corey Lynch and Andy Zeng and Oscar A. Ramirez and Ayzaan Wahid and Laura Downs and Adrian Wong and Johnny Lee and Igor Mordatch and Jonathan Tompson},
  title     = {Implicit Behavioral Cloning},
  booktitle = {Proceedings of the 5th Conference on Robot Learning},
  series    = {Proceedings of Machine Learning Research},
  volume    = {164},
  pages     = {158--168},
  publisher = {PMLR},
  year      = {2022},
  url       = {https://proceedings.mlr.press/v164/florence22a.html},
  eprint    = {2109.00137},
  archivePrefix = {arXiv}
}

@inproceedings{bc_z,
  author    = {Eric Jang and Alex Irpan and Mohi Khansari and Daniel Kappler and Frederik Ebert and Corey Lynch and Sergey Levine and Chelsea Finn},
  title     = {{BC-Z}: Zero-Shot Task Generalization with Robotic Imitation Learning},
  booktitle = {Proceedings of the 5th Conference on Robot Learning},
  series    = {Proceedings of Machine Learning Research},
  volume    = {164},
  year      = {2022},
  url       = {https://arxiv.org/abs/2202.02005}
}

@inproceedings{robomimic,
  author    = {Ajay Mandlekar and Danfei Xu and Josiah Wong and Soroush Nasiriany and Chen Wang and Rohun Kulkarni and Li Fei-Fei and Silvio Savarese and Yuke Zhu and Roberto Mart{\'i}n-Mart{\'i}n},
  title     = {What Matters in Learning from Offline Human Demonstrations for Robot Manipulation},
  booktitle = {Proceedings of the 5th Conference on Robot Learning},
  series    = {Proceedings of Machine Learning Research},
  volume    = {164},
  year      = {2022},
  url       = {https://arxiv.org/abs/2108.03298}
}

@misc{language_table,
  author        = {Corey Lynch and Ayzaan Wahid and Jonathan Tompson and Tianli Ding and James Betker and Robbie Baruch and Travis Armstrong and Pete Florence},
  title         = {Interactive Language: Talking to Robots in Real Time},
  year          = {2022},
  eprint        = {2210.06407},
  archivePrefix = {arXiv},
  primaryClass  = {cs.RO},
  url           = {https://arxiv.org/abs/2210.06407}
}

@misc{canonicalpolicy,
  author        = {Zhiyuan Zhang and Zhengtong Xu and Jai Nanda Lakamsani and Yu She},
  title         = {Canonical Policy: Learning Canonical 3D Representation for {SE(3)}-Equivariant Policy},
  year          = {2025},
  eprint        = {2505.18474},
  archivePrefix = {arXiv},
  primaryClass  = {cs.RO},
  doi           = {10.48550/arXiv.2505.18474},
  url           = {https://arxiv.org/abs/2505.18474}
}

@misc{sam2act,
  author        = {Haoquan Fang and Markus Grotz and Wilbert Pumacay and Yi Ru Wang and Dieter Fox and Ranjay Krishna and Jiafei Duan},
  title         = {{SAM2Act}: Integrating Visual Foundation Model with A Memory Architecture for Robotic Manipulation},
  year          = {2025},
  eprint        = {2501.18564},
  archivePrefix = {arXiv},
  primaryClass  = {cs.RO},
  doi           = {10.48550/arXiv.2501.18564},
  url           = {https://arxiv.org/abs/2501.18564}
}

@inproceedings{mikasa,
  author    = {Egor Cherepanov and Nikita Kachaev and Alexey K. Kovalev and Aleksandr I. Panov},
  title     = {Memory, Benchmark \& Robots: A Benchmark for Solving Complex Tasks with Reinforcement Learning},
  booktitle = {International Conference on Learning Representations},
  year      = {2026},
  url       = {https://openreview.net/forum?id=9cLPurIZMj}
}

@inproceedings{liberomem,
  author    = {Nhat Chung and Taisei Hanyu and Toan Nguyen and Huy Le and Frederick Bumgarner and Duy Minh Ho Nguyen and Khoa Vo and Kashu Yamazaki and Chase Rainwater and Tung Kieu and Anh Nguyen and Ngan Le},
  title     = {Rethinking Progression of Memory State in Robotic Manipulation: An Object-Centric Perspective},
  booktitle = {Proceedings of the AAAI Conference on Artificial Intelligence},
  year      = {2026},
  url       = {https://arxiv.org/abs/2511.11478},
  eprint    = {2511.11478},
  archivePrefix = {arXiv}
}

@misc{robomemarena,
  author        = {Huashuo Lei and Wenxuan Song and Huarui Zhang and Jieyuan Pei and Jiayi Chen and Haodong Yan and Han Zhao and Pengxiang Ding and Zhipeng Zhang and Lida Huang and Donglin Wang and Yan Wang and Haoang Li},
  title         = {RoboMemArena: A Comprehensive and Challenging Robotic Memory Benchmark},
  year          = {2026},
  eprint        = {2605.10921},
  archivePrefix = {arXiv},
  primaryClass  = {cs.RO},
  doi           = {10.48550/arXiv.2605.10921},
  url           = {https://arxiv.org/abs/2605.10921}
}

\end{document}